\documentclass[10pt,twocolumn,letterpaper]{article}

\usepackage[pagenumbers]{iccv} 

\usepackage{tikz}
\usepackage{wrapfig}
\usepackage{algorithm}
\usepackage{amsthm}
\usepackage{algpseudocode}
\usepackage[pagebackref,breaklinks,colorlinks,allcolors=iccvblue]{hyperref}
\usepackage{times}
\usepackage{epsfig}
\usepackage{graphicx}
\usepackage{amsmath}
\usepackage{amssymb}
\usepackage{booktabs}
\usepackage{appendix}
\usepackage{bbding}
\usepackage{makecell}

\usepackage{marvosym}
\usepackage{bm}
\usepackage{multirow}
\usepackage{colortbl} 
\newtheorem{theorem}{Theorem} 
\theoremstyle{definition}  

\definecolor{iccvblue}{rgb}{0.21,0.49,0.74}
\usepackage[accsupp]{axessibility}

\def\paperID{316} 
\def\confName{ICCV}
\def\confYear{2025}

\title{Mastering Collaborative Multi-modal Data Selection: \\A Focus on Informativeness, Uniqueness, and Representativeness}

\author{Qifan Yu$^1$\thanks{Equal contribution.} \quad Zhebei Shen$^1$\footnotemark[1]\quad  Zhongqi Yue$^2$\footnotemark[1] \quad Yang Wu$^3$ \quad Bosheng Qin$^1$ \quad Wenqiao Zhang$^1$ \quad \\\quad Yunfei Li$^3$ \quad Juncheng Li$^1$\textsuperscript{\Letter} \quad Siliang Tang$^1$\quad Yueting Zhuang$^1$\textsuperscript{\Letter} \\
\small $^1$Zhejiang University, $^2$Nanyang Technological University, $^3$Ant Group\\
{\tt\small \{yuqifan, shenzhebei, junchengli, siliang, yzhuang\}@zju.edu.cn}\\
{\tt\small nickyuezhongqi@gmail.com, wy306396@antgroup.com}\\\\}
\begin{document}
\maketitle
\renewcommand{\thefootnote}{\Letter}
\footnotetext[0]{Corresponding author.}
\begin{abstract}
 Instruction tuning fine-tunes pre-trained Multi-modal Large Language Models (MLLMs) to handle real-world tasks. However, the rapid expansion of visual instruction datasets introduces data redundancy, leading to excessive computational costs. We propose a collaborative framework, \textbf{DataTailor}, which leverages three key principles—--informativeness, uniqueness, and representativeness--—for effective data selection. We argue that a valuable sample should be informative of the task, non-redundant, and represent the sample distribution (\ie, not an outlier).
 We further propose practical ways to score against each principle, which automatically adapts to a given dataset without tedious hyperparameter tuning. Comprehensive experiments on various benchmarks demonstrate that DataTailor achieves 101.3\% of the performance of full-data fine-tuning with only 15\% of the data, significantly reducing computational costs while maintaining superior results. This exemplifies the ``Less is More" philosophy in MLLM development. The code and data is available in this \href{https://github.com/Yuqifan1117/DataTailor}{URL}.
\end{abstract}    
\section{Introduction}
\label{sec:intro}
The rapid development of Multi-modal Large Language Models~(MLLMs) has made promising progress on various multi-modal tasks~\cite{yu2023visually,alayrac2022flamingo,zhu2023minigpt}. 
A typical MLLM is developed through two main training stages: pre-training on vast image-text pairs and fine-tuning on task-specific multi-modal instructions.
Notably, the fine-tuning stage is critical for enhancing the instruction-following capabilities of MLLMs. Yet, this stage can become exceedingly time-consuming due to the large-scale but low-quality instruction data. Hence, the community is interested in fine-tuning data selection methods, such that an MLLM trained on the selected subset yields comparable or even better performance.

\begin{figure}[!t]
    \centering
    \includegraphics[width=1.\linewidth]{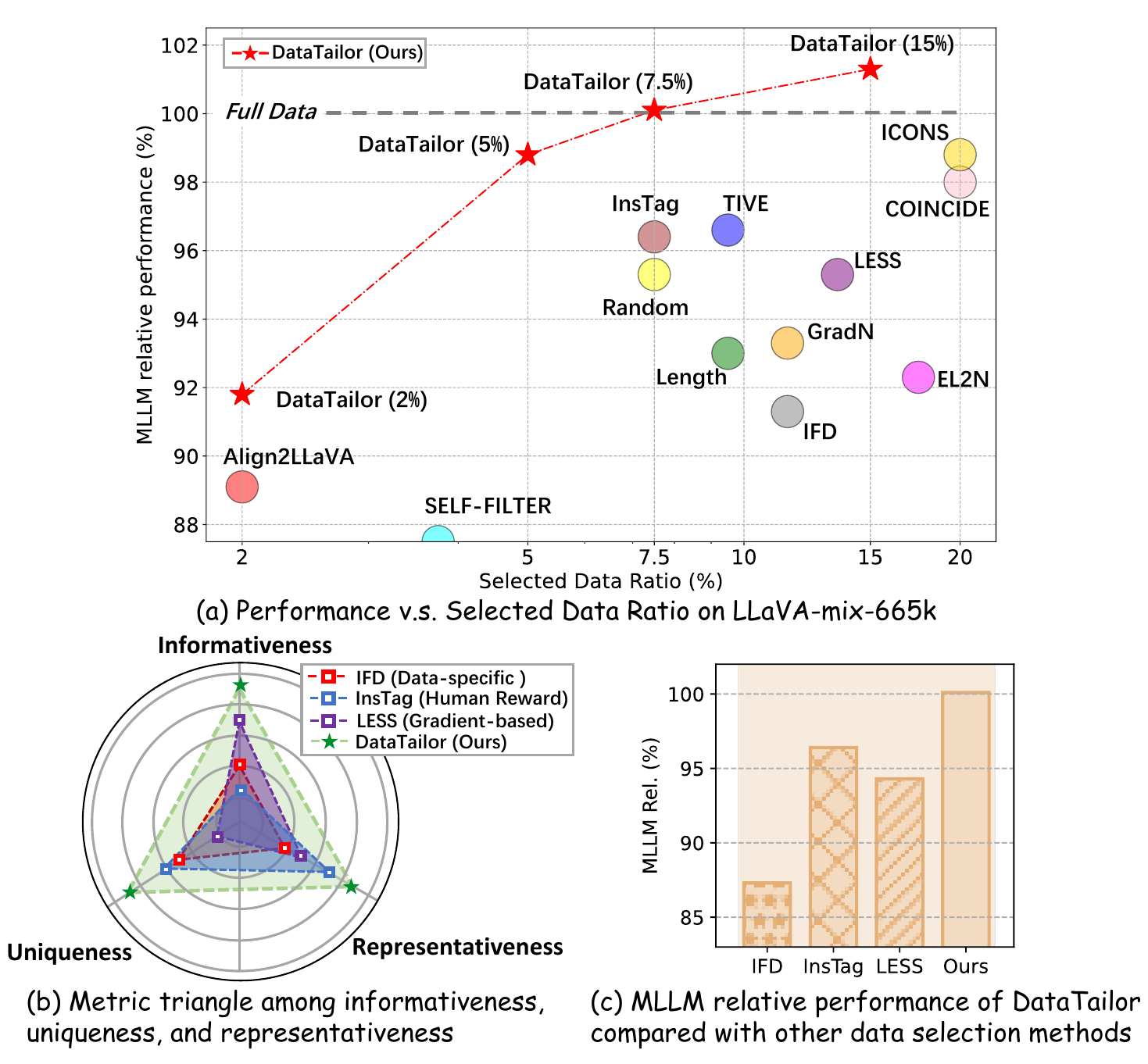}
    \vspace{-7mm}
    \caption{(a) The Performance v.s. Selected Data Ratio on LLaVA-mix-665k of DataTailor compared with SOTA data selection methods. (b) Metric triangle among informativeness, uniqueness, and representativeness when applying IFD~\cite{li2023quantity}~(data-specific methods), InsTag~\cite{lu2023instag}~(human-reward methods), LESS~\cite{xia2024less}~(gradient-based methods), and DataTailor. (c) The corresponding MLLM performance on LLaVA-mix-665k~\cite{liu2024improved} of different methods.} 
    \vspace{-6mm}
    \label{test}
\end{figure}
Existing MLLM data selection methods~\cite{dai2023instructblip, wu2023self, liu2024less, huang2024align} largely follow similar ideas from the NLP community~\cite{li2023quantity, lu2023instag, zhou2024lima, tan2024data, xia2024less}.
They can be divided into three main categories: (1) \textit{Data-specific methods}~\cite{li2023quantity, du2023mods, cao2023instruction} leverage vast hand-designed rules to select data in specific datasets, which are not flexible and robust for diverse tasks. (2) \textit{Human-reward methods}~\cite{zhou2024lima,lu2023instag} utilize human feedback to select data, which are both time-consuming and expensive. (3) \textit{Gradient-based methods}~\cite{tan2024data, xia2024less, lin2024data} select samples whose gradients are similar to the average training gradient of the dataset. 
However, they require additional training on downstream tasks, making total computation costly.

\begin{figure}[!t]
    \centering
    \includegraphics[width=1.\linewidth]{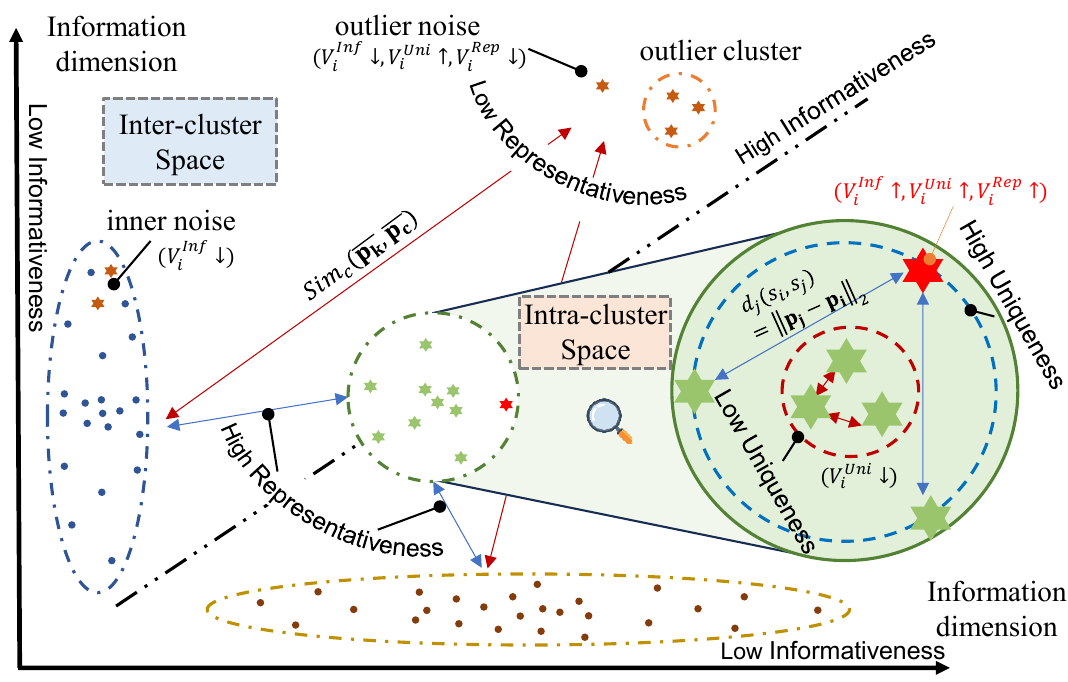}
    \vspace{-4mm}
    \caption{Illustration of the proposed method based on informativeness, uniqueness, and representativeness, where the x and y axes show information dimensions in latent space. The red star denotes a high-quality sample that satisfies the three principles.} 
    \vspace{-4mm}
    \label{attribute}
\end{figure}

To address these deficiencies, we build a systematic data selection method for MLLM called \textbf{DataTailor}. We evaluate each sample with three principles and select the most valuable samples, leading to state-of-the-art MLLM performance with a fraction of data (c.f. Fig.~\ref{test}). The three principles are:
(1) \textbf{Informativeness}: a valuable sample should be informative of the hard task, \eg, If the task is reasoning, describing the movement differences between skiing and ice skating is more informative and complex than simply describing someone skiing. In Fig.~\ref{attribute} where each axis (heuristically) represents an orthogonal dimension of task information, points along the diagonal carry more information about the task. (2) \textbf{Uniqueness}: A valuable sample should be distinct from others, offering unique insights rather than prevalent commonsense knowledge~(c.f. Fig.~\ref{attribute} near the blue dashed region in the intra-cluster space demonstrates high uniqueness).
(3) \textbf{Representativeness}: it should be a typical sample in the data distribution. This prevents selecting noisy outliers or mislabeled samples~(c.f. Fig.~\ref{attribute} the clusters connected by blue lines in the inter-cluster space exhibit high representativeness for the overall dataset).

We further propose a practical method to measure the value of each sample against each principle.
For informativeness, we take motivations from information theory~\cite{chen2019transferability, chen2023understanding}.
For each sample, we analyze the singular value distribution of its features and use the entropy of the singular values to determine if it is informative of the task.
To compute uniqueness and representativeness, we first cluster the samples based on their visual and textual features.
This allows efficient calculation for uniqueness, as we can simply measure the average distances of a sample to its neighbors in the same cluster, and mark those with a large distance as unique.
Then we find connected clusters and mark samples in those clusters as representative. Hence, noisy or mislabeled data from a far-away cluster can be filtered.
%

Moreover, as multi-modal samples exhibit varying structures and complexities across diverse tasks, we propose an adaptive weight to combine the values, which removes the need for expensive hyper-parameter tuning.
We also adaptively determine the proportion of selected data for each task by using the average largest singular value of samples in the task, which empirically reflects task difficulty and correlates with training robustness.
Combining these techniques, DataTailor synergizes the three principles for data selection and achieves an optimal balance between data volume and model performance~(as shown Fig.~\ref{test}(a) red line).

To our knowledge, we are the first to explore sample relationships between multi-modal instructions systematically. Through extensive experiments, we demonstrate that DataTailor exhibits significant effectiveness in data selection for MLLMs~(with less than 5\% data but achieving over 95\% performance). 
This effectiveness is further demonstrated through quantitative metrics designed for the three principles~(c.f. Fig.~\ref{test}(b)), proving that DataTailor improves data selection by addressing the essential aspects of each principle~(detailed analyses is in Sec.~\ref{exp4} and Appendix \textcolor{red}{D.1 \& D.2}).
In contrast, other methods lack a systematic evaluation, particularly of sample relationships, leading to weaknesses in uniqueness and representativeness and resulting in suboptimal MLLM performance~(c.f. Fig.~\ref{test}(c)). Remarkably, when DataTailor increases the data selection ratio, multi-modal data selection can even outperform full data fine-tuning~(achieve 101.3\% performance with 15\% data), truly exemplifying the concept of ``Less is More".
Overall, our main contributions are summarized as follows:
\begin{itemize}
    \item We identify three key principles~(\textit{i.e.}, informativeness, uniqueness, and representativeness) from a systematic perspective to master multi-modal data selection.
    \item We propose a unified framework, \textbf{DataTailor}, to adaptively integrate these principles for value evaluation to optimize multi-modal data selection in a collaborative way.
    \item 
    Extensive results show DataTailor's effectiveness in optimizing all three principles during selection and achieving new SOTA performance on various benchmarks.

\end{itemize}

\section{Related Work}
\label{related_work}
\subsection{Multi-modal Large Language Model}
With the outstanding performance of LLMs in zero-shot settings, early work combining LLMs with visual modalities has demonstrated impressive visual language comprehension abilities~\cite{guo2023images, yu2023visually, li2023fine, li2022fine, li2023variational, dai2023instructblip, li2023mimic}. Recently, more powerful MLLMs have emerged~\cite{zhu2023minigpt, pan2024auto, pan2024towards, ye2024mplug, liu2024improved, yu2025anyedit, yang2024qwen2, he2024efficient, bu2025limits}, which possess perceptual abilities for visual-language tasks and excellent reasoning abilities. Generally, the training process of MLLMs primarily includes two stages: the pre-training stage and the instruction tuning stage. Recent studies~\cite{yu2024hallucidoctor,liu2023mitigating, wu2024curriculum, ge2024worldgpt, gao2024fine, qiu2025step} have focused on the second stage to enhance instruction-following abilities. However, this stage is gradually facing inevitable computational overhead due to the growing volume of multi-modal data~\cite{liu2024less, qian2024momentor}. It is critical to identify a small subset of high-quality instructions to improve MLLM fine-tuning efficiency.
\subsection{Instruction-based Data Selection}
Although MLLMs have demonstrated remarkable performance, data redundancy is becoming apparent with the rapid growth of visual instruction datasets, similar to challenges in LLMs~\cite{zhou2024lima, cao2023instruction, everaert2023gio}. Previous LLM-based works mainly focus on using pre-defined specific rules~\cite{li2023quantity,cao2023instruction}, human feedback~\cite{lu2023instag}, or gradient-based approximation~\cite{ankner2024perplexed, xia2024less} to select high-quality data. INSTRUCTMINING~\cite{cao2023instruction} uses 9 indicators to fit specific rules for data selection. In contrast, DataTailor automatically assesses values without hand-designed rules, offering greater robustness and flexibility.
Moreover, these methods only focus on individual sample values while neglecting similar or noisy data when applied to more complex multi-modal instructions. For MLLM-based data selection,  TIVE~\cite{liu2024less} identifies redundancy in multi-modal instructions and selects valuable data at the task and instance level through gradient similarity, while ICONS~\cite{wu2024icons} extends this by integrating specialist influence estimation. However, they both require extra training on downstream tasks for data selection. SELF-FILTER~\cite{wu2023self} attaches evaluation models and updates its parameters during training to select high-value samples. InstructionGPT-4~\cite{wei2023instructiongpt} selects 200 instructions for MiniGPT4~\cite{zhu2023minigpt}, but it is unscalable for other settings. COINCIDE~\cite{lee2024concept} roughly clusters data by conceptual representations, while overlooking the value differences between clusters.
These methods largely follow prior approaches and overlook the complex relationships between samples, which limits their generalization.
To mitigate these limitations, we are the first to adopt a collaborative perspective for multi-modal data selection, focusing on three core principles: informativeness, uniqueness, and representativeness.
\section{Method}
\begin{figure*}[!t]
    \centering
    \includegraphics[width=1.\linewidth]{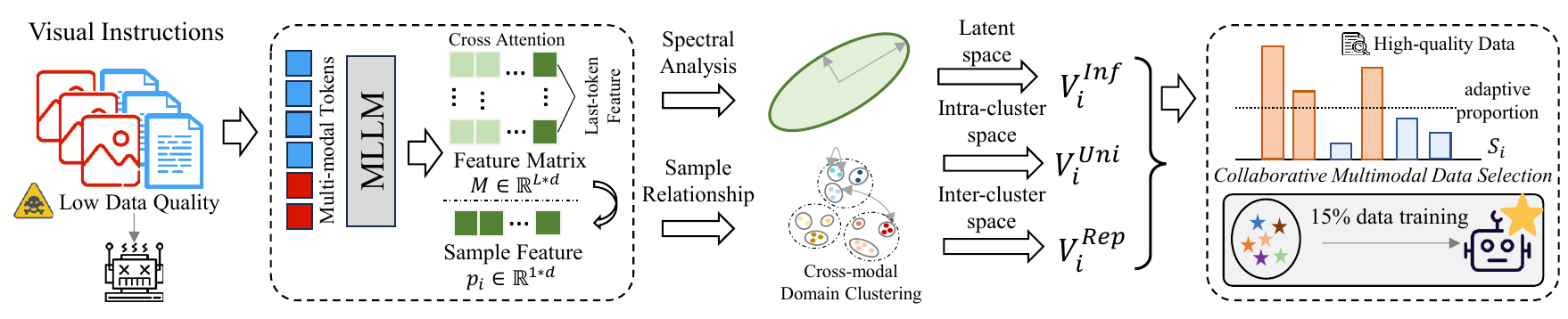}
    \vspace{-4mm}
    \caption{Overview of our proposed \textbf{DataTailor} for automatically selecting high-quality multi-modal data through the collaboration of three principled values (\textit{i.e.}, informative value, unique value, and representative value) from a systematic perspective.}
    
    \vspace{-4mm}
    \label{pipeline}
\end{figure*}
As illustrated in Figure~\ref{pipeline}, our DataTailor framework consists of four primary steps: (1)~The \textit{informative value} captures the information density in latent space, directly reflecting informativeness to enhance MLLM generalization.
(2)~The \textit{unique value} identifies distinct samples within the intra-cluster space, reflecting the uniqueness of sample relationships to reduce redundancy effectively. (3)~The \textit{representative value} captures samples that align closely with the overall dataset distribution in the inter-cluster space, ensuring representativeness and preventing compromise by noisy outliers. (4) Finally, DataTailor adaptively integrates these three values to enable collaborative multi-modal data selection.
Next, we will elaborate on the details of each step.
\subsection{Problem Formulation}
Given a visual input $X_v$ and a textual query $X_q$, the objective of the MLLM is to predict the correct answer $X_a$ for downstream tasks. We formulate multi-modal data selection as selecting fewest samples $S^*=\{s_1,...,s_k\}$ from the candidate dataset $S$ for MLLM training that achieves the best inference performance, where $s_i=(X_v, X_q, X_a)$ and $k$ is the total data selection proportion for $S$. Typically, the sample value is measured by the performance perturbations to the MLLM when removing the sample~\cite{koh2017understanding}, but this is not practical for large-scale datasets. 
Therefore, we propose three core principles~(\ie, informativeness, uniqueness, and representativeness) that practically identify the most valuable samples to improve MLLM's performance on downstream tasks. More concrete instantiations of each principle are in Appendix \textcolor{red}{D.2}. Then, we introduce each principle.

\subsection{Informative Value Estimation in Latent Space}
Although several approaches have been proposed for multi-modal data selection~\cite{wu2024icons, liu2024less, wei2023instructiongpt, wu2023self}, they are either infeasible at scale due to computational overhead or are limited in generalizability by pre-defined evaluation rules.
Previous studies~\cite{lin2024vila, liu2024visual, xia2024less} have shown that difficult samples contribute more to improving the performance of downstream tasks in MLLMs. Thus, DataTailor directly estimates samples' essential difficulty for multi-modal data selection.

However, it is non-trivial to quantify sample difficulty for multi-modal instructions due to varying information representations in different modalities. Drawing upon previous spectral
analysis~\cite{xue2022investigating, chen2019transferability, chen2023understanding}, we perform singular value decomposition~(SVD) on the unified feature matrix at the token level for image and text modalities to reflect the difficulty of samples, as it consists of the variations across different tokens in latent space. Formally, given a multi-modal sample $s_i$, we obtain its unified feature matrix $\mathbf{M_i}\in\mathbb{R}^{L_i*d}$ from the penultimate layer, where $L_i$ is the token length and $d$ is the feature dimension. We adopt this layer as the lower layers struggle to capture complex features, while the final layer suffers biases of pre-training datasets~\cite{liu2024visual}. Besides, the penultimate layer retains earlier layer information, providing richer features. Subsequently, we perform SVD on the matrix in latent space $\mathbf{M_i}=\mathbf{U_i}\hat{\mathbf{\Sigma_i}}\mathbf{V_i}^\top$ and its corresponding diagonal singular matrix is defined as follows:
\begin{equation}
    \hat{\mathbf{\Sigma_i}}=\{\sigma_1,...,\sigma_{L_i}\}
\end{equation}
where we assume $L_i \leq d$~\cite{liu2024improved} and all singular values $\{\sigma_j\}_{j=1}^{L_i}$ are listed in order. Building on this, we compute the entropy of normalized singular values as the informative value to assess the information density of each data:
\begin{equation}
    V_i^{Inf}=-\sum_{j=1}^{L_i}\frac{\sigma_j}{\sum_{k=1}^{L_i}\sigma_k}\log \frac{\sigma_j}{\sum_{k=1}^{L_i}\sigma_k}
    \label{eq2}
\end{equation}

Intuitively, simple samples contain redundant information in images or text, and only part of information is necessary to answer the query~(see Appendix \textcolor{red}{D.2} for examples). As a result, the columns of their feature matrices exhibit strong linear dependence, which drive down the smaller singular values, causing the top ones to be much bigger by comparison, leading lower singular value entropy. In contrast, more challenging samples are richer in information, with feature matrices that are closer to full rank, resulting in more uniform singular values and higher entropy. In this way, SVE can as a practical approximation to measure the difficulty of samples, enabling the selection of hard samples to enhance the performance of MLLM in downstream tasks.
\subsection{Unique Value in Intra-cluster Space}
To enrich the value of samples for multi-modal data selection, it is crucial to emphasize the unique values of samples to select unique samples from similar clusters. The unique value estimation in the intra-cluster space consists of two steps: \textit{cross-modal domain clustering} and \textit{unique value calculation}. Next, we introduce each step in detail.

\noindent\textbf{Cross-modal Domain Clustering.} 
To efficiently distinguish unique samples, we propose an optimized Cross-modal Domain Clustering that aggregate sufficiently similar samples into cluster in each task. This process begins with quantifying the $\ell_2$-norm distance between samples for cluster variance computation. 
It then utilizes the Ward criterion~\cite{mullner2011modern} to merge clusters with the minimum increase in sum of squared errors (SSE) progressively, computed as:
\begin{equation}
    \Delta \text{SSE} = \frac{n_A\cdot n_B}{n_A + n_B}\cdot \|\boldsymbol{\mu}_A - \boldsymbol{\mu}_B\|_2
\end{equation}
where $n_A,n_B$ denote cluster cardinalities and $\boldsymbol{\mu}_A,\boldsymbol{\mu}_B$ represent cluster centroids. The clustering terminates when the $\Delta \text{SSE}$ exceeds the threshold $\lambda\Delta \text{SSE}_{\text{max}}$, where $\Delta \text{SSE}_{\text{max}}$ denotes the maximum increase in SSE. Here, $\lambda$ controls the number of clusters and is set to 0.1 by default to match the total data selection proportion for capturing sample relationships.
Moreover, we use parallel computation and memory optimization for efficiency. 
More implementation details and analyses are shown in Appendix \textcolor{red}{B.3}.

\noindent\textbf{Unique Value Calculation.} 
To quantify the uniqueness of samples, we focus on identifying discriminative samples in the intra-cluster space that contribute uniquely to training. 
The key intuition is that unique samples exhibit larger distance from similar samples~(as shown in Fig.~\ref{attribute}) to contain more information in the cluster for generalization.
Therefore, we introduce a distance coefficient to assign high unique values to these distinctive instructions based on their distance from surrounding samples as follows:
 \begin{equation}
V_i^{Uni}=\sum_{s_j\in \mathbf{C},j\neq i}\Vert\mathbf{p_j}-\mathbf{p_i}\Vert_2\cdot \frac{V_j^{Inf}}{\sum_{k\in\mathbf{C}}V_k^{Inf}}
\label{eq8}
\end{equation}
where $\Vert\mathbf{p_j}-\mathbf{p_i}\Vert_2$ is the Euclidean distance of two samples $s_i,s_j$ in the cluster $\mathbf{C}$ in latent space. 
We assign higher reference weights to samples with greater informative value when computing their unique and representative values, reflecting their stronger influence on other samples. In this manner, these challenging instructions are more likely to be selected due to their enhanced unique values. 
\subsection{Representative Value in Inter-cluster Space}
Empirically, evaluating samples solely with their informativeness and uniqueness within clusters may introduce outlier noisy or mislabeled data. These samples, despite higher uniqueness, have weaker associations with other clusters, limiting their ability to represent the overall dataset.
Therefore, we introduce an inter-cluster representative coefficient to measure relationships across clusters, ensuring that selected samples capture representative features from the overall dataset, thereby avoiding the selection of noisy data:
\begin{equation}
    \tau_i^c=\frac{1}{K-1}\sum_{k\neq c}^K\exp({\operatorname{sim}}(\mathbf{\overline{p_k}},\mathbf{\overline{p_c}}))
\end{equation}
where $\mathbf{\overline{p_c}}$ is the average sample feature in the target cluster $\mathbf{C}$ that contains sample $s_i$, $\{\mathbf{\overline{p_k}}\}_{k\neq c}^K$ is the average sample feature of other clusters and $K$ is the number of clusters in each task. We use the feature of the last token to represent the sample feature as it aggregates all visual and textual features by cross-attention. Here, $\operatorname{sim}(\cdot, \cdot)$ is the cosine similarity and $\exp(\cdot)$ is used to amplify the effect of clusters. Based on this coefficient, we then assign the weighted representative value to the instruction $s_i$ as follows:
\begin{equation}
V_i^{Rep}=\tau^c_i\cdot \frac{V_i^{Inf}}{\sum_{k\in\mathbf{C}}V_k^{Inf}}
\label{eq5}
\end{equation}
In this way, the representative value uses the association coefficient to ensure that selected samples align with the overall distribution. When the value is high, it indicates that the selected samples can effectively represent other samples, reducing the impact of noisy data and enhancing the overall representativeness in conjunction with uniqueness. More implementation details are shown in Appendix \textcolor{red}{C.1}.

\subsection{Adaptively Collaborative Data Selection}
\begin{figure}[!t]
    \centering
    \includegraphics[width=0.85\linewidth]{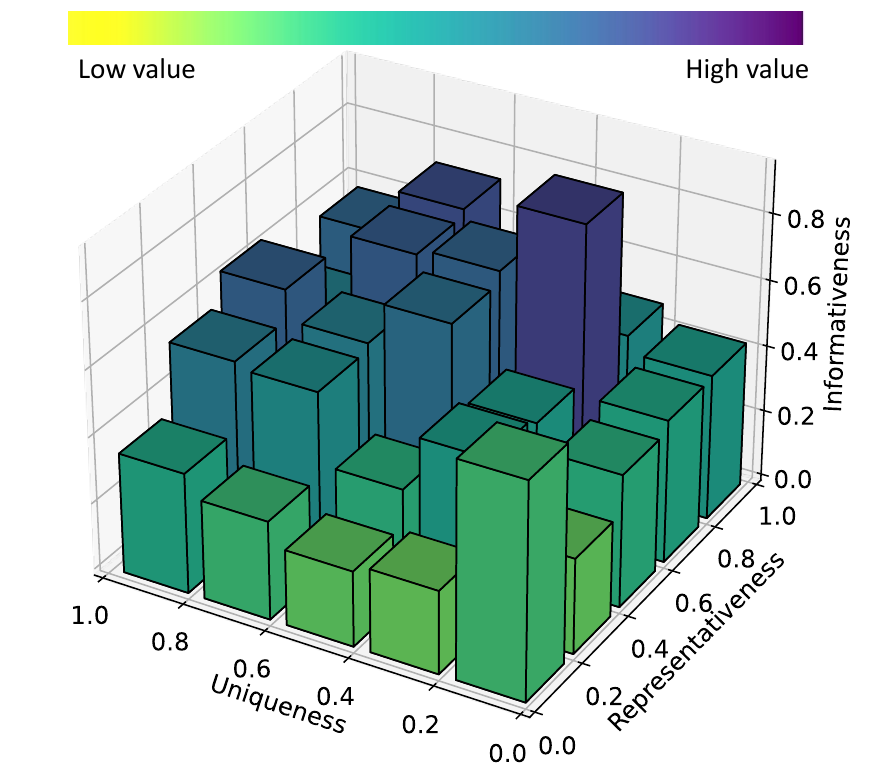}
    \vspace{-0.2cm}
    \caption{Visualization of the collaboration among informativeness, uniqueness, and representativeness for data selection, where each bar represents a subset defined by a specific value interval.}
    \label{collaboration}
    \vspace{-0.5cm}
\end{figure}

Although we obtain multi-scale values from three complementary perspectives, combining them to select ideal samples is challenging since these multi-modal samples exhibit varying instruction rounds. Multi-IF~\cite{he2024multi} points out that multi-turn instructions prioritize informative value due to their weak interrelationships while single-turn ones emphasize unique and representative value due to their limited information. Inspired by this, we introduce an influence factor based on the number of response rounds of each multi-modal instruction for adaptively collaborative data selection to enable adaptive, collaborative data selection, enhancing the synergy among these three values as follows:
\begin{equation}
V_i=\frac{r_i}{r_i+2}\cdot{V}_i^{Inf}+\frac{1}{r_i+2}\cdot({V}_i^{Uni}+{V}_i^{Rep})
\end{equation}
where $r_i$ denotes the conversation round of each multi-modal instruction. We employ a 1:1 ratio to simplify the balance between uniqueness and representativeness values, as their trade-off remains stable. With this synergistic value for data selection, we can identify informative and unique instructions while adequately representative~(c.f., Fig.~\ref{collaboration}). 
We also show detailed demonstration of DataTailor addressing each of three principles in Sec.~\ref{exp4} and Appendix \textcolor{red}{D.1}. 

In addition, we observe that standardizing the data selection proportion across all tasks limits selection diversity due to differences in task difficulty. 
Since spectral analysis shows that samples with the higher largest singular value exhibit less directional diversity for generalization, they require more selected data to improve the model's training robustness.
Thus, we propose adaptive data selection proportion $k_p$ for each task $S_p$ based on largest singular values, which is computed as follows:
\begin{equation}
k_p=\frac{x_p^2\cdot |S_p|}{\sum_{q}x_q^2\cdot |S_q|}\cdot k,\quad x_p = avg({\frac{\sigma_{\text{max}}}{\sum_{j=1}^{L_i}\sigma_j}})
\end{equation}
where $x_q$ is the average ratio of the largest singular value $\sigma_{\text{max}}$ to the sum of all singular values for all samples in the task $S_q$, $|S_q|$ is the number of samples in the task, and $k$ is the whole data selection proportion.
Through collaborative value assessment with task-adaptive proportions, DataTailor promotes more diversity during MLLM data selection. More details and analyses are shown in Appendix \textcolor{red}{B.2}.

 \begin{table*}[!t]
\resizebox{0.99\textwidth}{!}{
\begin{tabular}{lcccccccc|ccccc|c|c}
\toprule
\multirow{2}{*}{Methods} & \multirow{2}{*}{Valid Data}& \multicolumn{7}{c}{MLLM Benchmarks} & \multicolumn{5}{c}{VQA Benchmarks} & \multicolumn{1}{c}{Captioning} \\
& & \multicolumn{1}{c}{MME-P $\uparrow$} & \multicolumn{1}{c}{MME-C $\uparrow$} & \multicolumn{1}{c}{SEED-I $\uparrow$} & \multicolumn{1}{c}{POPE $\uparrow$}&  \multicolumn{1}{c}{MM-Vet $\uparrow$}&  \multicolumn{1}{c}{LLaVA-Wild $\uparrow$}&  \multicolumn{1}{c}{MMMU(val) $\uparrow$}& \multicolumn{1}{c}{VizWiz $\uparrow$}&\multicolumn{1}{c}{SciQA $\uparrow$} & \multicolumn{1}{c}{GQA $\uparrow$} & \multicolumn{1}{c}{VQA-v2$\uparrow$}& \multicolumn{1}{c}{TextVQA $\uparrow$} & \multicolumn{1}{c}{NoCaps (val) $\uparrow$} & Rel.\\\hline

&\multicolumn{14}{c}{\textbf{\small MiniGPT4-Instruction}}  & \multicolumn{1}{r}{\scriptsize MiniGPT4-7B}\\\hline
MiniGPT4-7B& 3.4k & 717.4 & 259.6 &23.8&68.3& 19.0 & 25.1 & 23.4 & 36.0&36.3&32.2&32.1&21.4&111.5&100.0\%\\\hline
\quad Random&0.2k&698.4&227.7&25.1&69.7& 17.2 & 27.9 & 19.1 & 18.3&34.0&19.2&33.2&17.2&105.1&89.1\%\\
\quad Length&0.2k&683.4&209.6&26.7&69.8& 18.3 & 26.7 & 17.4& 29.9&35.6&32.5&33.7&17.4&106.5&94.7\%\\
\quad EL2N~\cite{paul2021deep}&0.2k&668.6&207.6&26.5&72.0& 17.4 & 27.3 & 23.1 & 41.9&36.1&32.9&36.3&23.7&108.3&102.2\%
\\\hline
\quad IFD~\cite{li2023quantity} &0.2k&678.6&213.8&29.1&47.4& 18.9 & 28.4 & 20.6 & 42.7&38.1&28.3&36.0&23.4&106.6&99.8\%\\
\quad InsTag~\cite{lu2023instag} &0.2k&715.6&237.9&26.8&70.4 & 15.8 & 28.1 & 21.7 &40.0&38.1&30.1&34.5&22.2&105.9&100.8\%\\
\quad LESS~\cite{xia2024less} &0.2k&698.5&191.4&22.4&71.8 & 19.8 & 25.8 & 22.2 &38.4&35.4&26.0&34.4&16.6 &109.7&95.4\%\\\hline
\quad InstructionGPT-4~\cite{wei2023instructiongpt}&0.2k&716.9&229.6&17.4&71.6& 21.3 & 25.3 & 14.9 & 29.9&35.1&26.8&34.8&22.1&106.8&93.3\%\\
\quad SELF-FILTER~\cite{wu2023self}  & 0.5k &438.7&128.6&21.7&71.4& 19.4 & 24.3 & 20.4 & 41.3&35.7&30.4& 35.0& 22.0&105.6&92.8\%\\
\quad TIVE~\cite{liu2024less}&0.2k&707.0&200.9&23.6&72.3& 17.5 & 25.8 & 18.2 & 31.4&33.8&26.4&35.1&17.5&108.9&92.7\%\\\hline
\cellcolor{gray!30}{\quad DataTailor~(Ours)}&\cellcolor{gray!30}0.2k&\cellcolor{gray!30}720.6&\cellcolor{gray!30}263.9&\cellcolor{gray!30}27.3&\cellcolor{gray!30}69.8&\cellcolor{gray!30}21.4&\cellcolor{gray!30}28.4&\cellcolor{gray!30}23.6&\cellcolor{gray!30}40.8&\cellcolor{gray!30}37.7&\cellcolor{gray!30}30.7&\cellcolor{gray!30}34.7&\cellcolor{gray!30}21.0&\cellcolor{gray!30}106.9&\cellcolor{gray!30}104.6\% \\ \hline
&\multicolumn{14}{c}{\textbf{\small LLaVA-1.5-mix-665k}}  & \multicolumn{1}{r}{\scriptsize LLaVA-7B}\\\hline
LLaVA-v1.5-7B~(LoRA) & 665k & 1476.9 & 267.9 & 67.4 &86.4& 30.9 & 67.9 & 32.8 & 47.8&70.0&63.0&79.1&58.2&106.5&100.0\% \\\hline
\quad  Random&50k&1387.5&287.5&59.7&85.7& 29.5 & 64.5 & 32.2 & 42.3&70.0&55.0&73.7&53.1&107.7&95.3\% \\
\quad Length&50k&1357.0&265.7&47.0&82.6& 29.7 & 67.6 & 33.9 & 49.2&60.9&55.5&70.7&45.2&88.2&91.0\% \\
\quad EL2N~\cite{paul2021deep}&50k&1077.3&252.5&59.3&80.8 & 21.1 & 40.1 & 33.6 &44.4&\underline{71.0}&41.7&61.0&41.7&86.9&82.3\%\\
\quad GradN~\cite{paul2021deep}&50k&1275.4&303.6&58.3&75.7& 24.8 & 68.2 & 32.4 & 37.8&70.9&44.9&64.0&46.0&101.9&89.3\% \\\hline
\quad IFD~\cite{li2023quantity}&50k&1113.4&301.8&55.1&76.7& 27.6 & 63.1 & 33.0 & 48.7&48.2&41.9&64.2&43.6&106.8&87.3\% \\
\quad InsTag~\cite{lu2023instag} & 50k & 1317.1 & \underline{345.0} &57.4&82.1 & 29.6 & 68.1 & \textbf{34.0} &47.4&69.3&52.5&63.2&53.3 & \underline{108.3} & 96.4\% \\
\quad LESS~\cite{xia2024less} & 50k & 1344.8 & 281.8 & 61.2 & 79.4 & 28.3 & 65.5 & 33.0 & 44.4 & \underline{71.0} & 53.4 & 71.8 &  52.0&106.2 & 94.3\% \\\hline
\quad SELF-FILTER~\cite{wu2023self} & 25k & 955.7 & 262.5 & 47.5 &76.0& 26.6 & 60.3 & 30.6 & 40.8&59.4&3.6&2.1&5.6&82.3 & 65.8\% \\ 
\quad TIVE~\cite{liu2024less}&50k&1334.8&248.6&\underline{62.2}&85.9& 30.2 & 67.9 & 33.1 & 45.1& \textbf{71.4} & 56.2 & 73.8 & 51.1 &96.0 & 94.6\% \\
\quad  
COINCIDE~\cite{lee2024concept}&133k&\textbf{1495.6}&-&-&\underline{86.1} & - & 67.3 & - &46.8& 69.2 & 59.8 & \underline{76.5} & \underline{55.6} & - & 98.0\% \\
\quad 
ICONS~\cite{wu2024icons}&133k&\underline{1485.7}&-&-&\textbf{87.5}& 29.7 & 66.1 & - & \textbf{50.1}& 70.8 & \textbf{60.7} & 76.3 & \underline{55.6} & - & 98.8\% \\\hline
\cellcolor{gray!30}{\quad DataTailor~(Ours)}&\cellcolor{gray!30}50k&\cellcolor{gray!30}1461.2&\cellcolor{gray!30}\textbf{362.5}&\cellcolor{gray!30}61.7&\cellcolor{gray!30}82.1&\cellcolor{gray!30}\underline{30.4}&\cellcolor{gray!30}\underline{69.3}&\cellcolor{gray!30}\underline{33.9}&\cellcolor{gray!30}46.3&\cellcolor{gray!30}70.9&\cellcolor{gray!30}57.7& \cellcolor{gray!30}75.0 &\cellcolor{gray!30}53.1 &\cellcolor{gray!30}107.2 & \cellcolor{gray!30}\underline{100.1\%}
\\
\cellcolor{gray!30}{\quad DataTailor~(Ours)}&\cellcolor{gray!30}100k&\cellcolor{gray!30}1476.2&\cellcolor{gray!30}319.3&\cellcolor{gray!30}\textbf{63.6}&\cellcolor{gray!30}85.3&\cellcolor{gray!30}\textbf{31.8}&\cellcolor{gray!30}\textbf{71.1}&\cellcolor{gray!30}33.2&\cellcolor{gray!30}\underline{49.5}&\cellcolor{gray!30}\underline{71.0}& \cellcolor{gray!30}\underline{60.5} & \cellcolor{gray!30}\textbf{76.7} & \cellcolor{gray!30}\textbf{55.7} & \cellcolor{gray!30}\textbf{108.7} & \cellcolor{gray!30}\textbf{101.3\%} \\
\bottomrule
\end{tabular}
}
\caption{Comprehensive comparison between DataTailor and other baselines for multi-modal data selection on MLLM and downstream general benchmarks. Our results are shown in the gray block. Due to limited resources, we all use the LoRA model for fair comparisons.}
\label{tab1}
\vspace{-0.4cm}
\end{table*}

\section{Experiments}
We first evaluate DataTailor on the standard data selection of MLLM on various benchmarks~(\S~\ref{exp2}). Additionally, we examine its transferability to other datasets~(\S~\ref{exp3}) and conduct an in-depth analysis~(\S~\ref{exp4}) for further evaluation.

\subsection{Experimental Setup}
\noindent\textbf{Multi-modal Instruction Data \& Backbone.} As ideal data selections should be adaptable to diverse MLLM instruction datasets, we integrate DataTailor with two widely-used datasets to conduct experiments for its effectiveness evaluation: 1) MiniGPT4-Instruction~\cite{zhu2023minigpt} includes about 3.5K instances refined by ChatGPT from detailed descriptions.
2) LLaVA-1.5-mix-665k~\cite{liu2024improved} is a wider collection with 665K instructions, which encompass a wide range of task categories, including dialogue-based Q\&A pairs, multiple-choice short Q\&A, detailed descriptions, and text-only reasoning tasks. For the general setting, we conduct experiments on MiniGPT-4-7B and  LLaVA-v1.5-7B. 



\noindent\textbf{Benchmarks.} We assess our methods using a mix of MLLM-specific benchmarks and more general downstream tasks.
Aligned with SOTA MLLM methods, we include MLLM benchmarks for comprehensiveness: MME~\cite{fu2024mmecomprehensiveevaluationbenchmark} is used to evaluate MLLM’s ability of perception and cognition; SEED-Bench~\cite{li2024seed} involves multi-modal tasks across 12 perspectives with the assistance of GPT-4; POPE~\cite{li2023evaluating} mainly evaluates the MLLM’s hallucination problems; MM-Vet~\cite{yu2023mm} and LLaVA-Wild~\cite{liu2024visual} assess the model's open-ended conversational capabilities; The MMMU~\cite{yue2024mmmu} consists of more challenging scientific problems to assess reasoning ability.
For general VQA tasks, VizWiz~\cite{gurari2018vizwiz} and ScienceQA~\cite{saikh2022scienceqa} contain unseen visual queries and multiple-choice questions to evaluate the zero-shot generalization of MLLMs. VQA-v2~\cite{antol2015vqa} and GQA~\cite{hudson2019gqa} access the model’s visual perception abilities with open-ended questions while TextVQA~\cite{singh2019towards} focuses on text-rich questions. For captioning, we transfer MLLMs to the NoCaps~\cite{agrawal2019nocaps} validation set. We also show the amount of valid data selected to demonstrate the effectiveness of data selection. Note that \textbf{Rel.} in all tables represents the relative boost on all benchmarks

\noindent\textbf{Baselines.} We use the following baselines:
1) \textbf{Traditional data selection}: it includes traditional random selection; length-based selection;
GradN~\cite{paul2021deep} and EL2N~\cite{paul2021deep} use the L2-norm of the gradient and the error vector for selection, respectively. 
2) \textbf{LLM data selection}: it directly transfers the selection methods from LLMs to MLLMs, including data-specific methods~\cite{cao2023instruction}, human-reward methods~\cite{lu2023instag}, and gradient-based method~\cite{xia2024less}. 3) \textbf{MLLM-specialized selection}: it involves methods specifically designed for data selection in MLLM, including InstructionGPT-4~\cite{wei2023instructiongpt}, SELF-FILTER~\cite{wu2023self}, TIVE~\cite{liu2024less}, COINCIDE~\cite{lee2024concept}, ICONS~\cite{wu2024icons}.


\subsection{Main Results on Multi-modal Data Selection}
\label{exp2}
We report the results of our DataTailor and other diverse data selection methods for the MiniGPT4 and LLaVA shown in Table~\ref{tab1}.
Based on the observation of experimental results, we have summarized the following conclusions: 

\textbf{Multi-modal instruction data suffers from serious redundancy, resulting in overall poor data quality.} We can observe that in most benchmarks, even randomly selecting a small amount of instruction data does not result in a performance drop. 
Moreover, in some cases, simply selecting part of the data outperforms utilizing the entire dataset~(33.2 v.s. 32.1 of VQA-v2 on MiniGPT-4), suggesting that excessive low-quality data hinder several MLLMs' capabilities on the contrary. Qualitatively, as shown in Fig.~\ref{test}, 
most methods achieve 80\% performance with less than 20\% of the data. 
This confirms our analysis of the data redundancy in multi-modal datasets and the necessity of data selection.

\textbf{For LLM data selection approaches~(
\textit{i.e.}, IFD~\cite{li2023quantity}, InsTag~\cite{lu2023instag}, and LESS~\cite{xia2024less}), the performances across several benchmarks overall remain unsatisfactory.} 
Although these baselines explicitly distinguish tasks and features, they still underperform DataTailor by 12.8\%, 3.7\%, and 5.8\% on overall relative performance due to their rough information modeling and disregard for sample relationships. Moreover, all LLM data selection methods demonstrate severe shortcomings in representativeness, which leads to a decline in the performance of general VQA tasks~(average 49.6 in TextVQA and 49.3 in GQA). In contrast, we observe that DataTailor obtains overall improvement on various tasks. For an intuitive illustration, we visually compare the metrics of three principles of DataTailor and those LLM data selection methods, as shown in Figure~\ref{test}(b). 
Similarly, those LLM data selection methods exhibit deficiencies in at least one dimension, whereas DataTailor consistently achieves promising results.
This result demonstrates DataTailor's capability to effectively promote these principles for multi-modal data selection, rather than roughly selecting samples based on individual values.

\textbf{Our DataTailor can be flexibly applied to different MLLMs.} We incorporate our DataTailor into the two most popular MLLMs for evaluation. Despite the diversity in data and model structures, our DataTailor consistently improves relative performance of data selection across all benchmarks compared to random selection~(e.g., +15.5\% on MiniGPT4-7B and +4.8\% on LLaVA-v1.5-7B).  
Notably, despite comprising only 7.5\% data, DataTailor consistently outperforms full fine-tuning on the challenging MMMU benchmark~(23.6 v.s. 23.4 on MiniGPT4-7B and 33.9 v.s. 32.8 on LLaVA-7B).
These results indicate that our proposed method consistently addresses data redundancy across different datasets and architectures.

\textbf{Compared with MLLM-specialized selection methods, DataTailor exceeds SOTA for overall benchmarks.} Specifically, DataTailor achieves an average of 100.1\% relative performance, outperforming TIVE~\cite{liu2024less} (94.6\%) and the latest ICONS~\cite{wu2024icons} (98.8\%). 
Notably, when increasing data selection ratio to 15\%, DataTailor surpasses the performance of the full tuning of LLaVA-v1.5-7B~(103.0\% for MLLM benchmarks and 101.3\% for total benchmarks).
This indicates that a small amount of high-quality data is more crucial than a large volume of low-quality data for enhancing MLLMs, which truly exemplifies ``Less is More".

\begin{table}[!t]
\vspace{-0.2cm}
\resizebox{0.48\textwidth}{!}{
\begin{tabular}{lcccccc}
\hline
\toprule
\multicolumn{1}{c}{Methods} & Feature Backbone& \multicolumn{1}{c}{MME$\uparrow$} & \multicolumn{1}{c}{SEED-I $\uparrow$} & \multicolumn{1}{c}{MMMU(val)$\uparrow$}&\multicolumn{1}{c}{SciQA $\uparrow$}  & Rel.\\\hline

\multicolumn{2}{l}{mPLUG-264k~\cite{ye2024mplug}}\\
\quad Full Data (100\%)&-&1243.4&34.3&26.9&41.1&100.0\%\\
\quad Random (5\%)&-&1183.2&33.9&26.6&40.6&97.9\%\\
\quad TIVE~\cite{liu2024less} (5\%)&LLaVA-vicuna-7B
&1177.1&34.0&26.9&40.2&97.9\%\\\hline
\multirow{2}{*}{\quad DataTailor (5\%)}&LLaVA-vicuna-7B&\textbf{1260.0}&\textbf{34.5}&27.8&\textbf{41.9}&\textbf{101.8\%}\\
&mPLUG-OWL-7B&1217.2&34.3&\textbf{28.0}&41.3&100.6\%\\\hline
\multicolumn{2}{l}{Bunny-695k~\cite{he2024efficient}} \\
\quad Full Data (100\%)&-&1778.1&70.7&38.7&70.9&100.0\%\\
\quad Random (5\%)&-&1578.8&64.4&36.7&70.1&93.4\%\\
\quad TIVE~\cite{liu2024less} (5\%)&LLaVA-vicuna-7B&1542.4&61.7&34.7&68.4&90.0\%\\\hline
\multirow{2}{*}{\quad DataTailor (5\%)}&LLaVA-vicuna-7B&1582.8&\textbf{65.3}&37.1&\textbf{75.8}&\textbf{96.0\%}\\
&Bunny-Phi-3B&\textbf{1599.8}&63.3&\textbf{37.3}&74.4&95.2\%\\
\bottomrule
\end{tabular}
}

\caption{Transferability analysis of multi-modal data selection.}
\label{Transferability}
\vspace{-0.4cm}
\end{table} 
\subsection{Transferability of Multi-modal Data Selection}
\label{exp3}
Previous MLLM-specialized methods~\cite{liu2024less, xia2024less} relied on the same MLLMs as in the training phase for data selection. The transferability analysis aims to investigate whether multi-modal data selected by surrogate models can be efficiently transferred into the target MLLM.
Here, we use LLaVA-v1.5-7B as the surrogate model to select valuable data for the target MLLMs mPLUG-Owl-7B~\cite{ye2024mplug} and Bunny-3B~\cite{he2024efficient}, whose corresponding target datasets are mPLUG-264k and Bunny-695k, respectively. Moreover, we explore DataTailor’s robustness among other feature backbones with different structures and parameters. Table~\ref{Transferability} presents the results of our DataTailor and other baselines


We observe that, (a) despite inconsistencies between the data selection model and the target MLLMs, DataTailor still consistently achieves over 95\% relative performance of full fine-tuning with only 5\% data~(101.8\% in mPLUG-264k and 96.0\% in Bunny-695k). This demonstrates the powerful transferability of DataTailor and its potential for surrogate data selection. In contrast, TIVE~\cite{liu2024less} performs similarly to or worse than random selection~(\eg, 90.0\% v.s. 93.4\% in Bunny-695k) though it outperforms it in Tab.~\ref{tab1}. This discrepancy may stem from TIVE's strong correlation with training gradients of the specific model. (b) The stable performance transition (nearly 1\%) indicates that DataTailor emphasizes the inherent value of samples rather than relying on the model features as in previous feature-based baselines, which highlights its transferability for data selection.


\begin{table}[!t]
\vspace{-0.2cm}
\setlength\tabcolsep{11pt}
\resizebox{0.48\textwidth}{!}{
\begin{tabular}{cl|ccc|ccc|c}
\hline
\toprule
&\multicolumn{1}{c}{\multirow{2}{*}{Methods}} & 
\multicolumn{3}{c}{Principled Values} &\multicolumn{3}{c}{Benchmarks}   \\ 
&  & $V_i^{Inf}$& $V_i^{Uni}$& $V_i^{Rep}$ & MME $\uparrow$ & MMMU(val) $\uparrow$ & SciQA $\uparrow$ & Rel.\\ \hline
 1 &Full Data & - & - & - & 1744.8 & 32.8  & 70.0 &100.0\% \\ \hline
 2&Random & \XSolidBrush & \XSolidBrush & \XSolidBrush & 1675.0 & 32.2 & 70.0 & 95.3\%  \\ 
 3&\quad +$V_i^{Inf}$  & \Checkmark & \XSolidBrush & \XSolidBrush & 1759.3 & \textbf{34.9} & 70.2 & 98.0\% \\ 
 4&\quad +$V_i^{Uni}$  & \XSolidBrush & \Checkmark & \XSolidBrush & 1716.2 & 33.5 & 69.8 & 97.3\% \\
 5&\quad +$V_i^{Rep}$ & \XSolidBrush & \XSolidBrush & \Checkmark & 1771.4 & 33.8 & 68.5 & 97.5\% \\\hline
 6&\textbf{DataTailor}  & \Checkmark & \Checkmark & \Checkmark & \textbf{1823.7} & 33.9 & \textbf{70.9} & \textbf{100.1\%}  \\
 7&\quad w/o adaptive collaboration 
 & \Checkmark & \Checkmark & \Checkmark & 1770.2 & 34.0 & 70.2  &  98.8\% \\
 8&\quad w/o adaptive proportion
 & \Checkmark & \Checkmark & \Checkmark & 1753.4 & 33.3 &  70.1 &  97.7\% \\
 9&\quad w/o adaptive collaboration \& proportion
 & \Checkmark & \Checkmark & \Checkmark & 1730.0 & 32.4 &  69.3 & 97.2\% \\\hline
\end{tabular}}
\caption{Ablation study of each module in DataDailor. All experiments are with 7.5 \% selection proportion on LLaVA-mix-665k and Rel. denotes the average boost on all 13 benchmarks.}
\label{ablation}
\vspace{-0.5cm}
\end{table}

\subsection{In-depth Analysis}
\label{exp4}
\noindent\textbf{Analysis of Instruction Selection Factors.} 
To investigate our DataTailor deeply, we study the ablation variants of different factors in Table~\ref{ablation}.
Specifically, we analyze their independence using the following ablation strategy: 
1) +$V_i^{Inf}$: we only use the informative value. 2) +$V_i^{Uni}$: we only include the unique value. 3) +$V_i^{Rep}$: we only consider the representative value. 4) w/o adaptive collaboration \& proportion: we gradually remove adaptive weights for collaboration and adaptive proportions for selection.
Note that we adopt adaptive proportion in all Row 2-5 for fair comparison.
The results of Row 3 indicate that informative value is the most crucial principle. 
Also, Row 4 and 5 suggest the importance of unique and representative values, as unique values support MLLMs' discriminative capabilities, while representative values enhance their generative capabilities. Furthermore, decreasing performance in rows 7-9 suggests that the adaptively collaborative strategy effectively ensures diversity between tasks in multi-modal data selection.

\noindent\textbf{Instantiation of Three Principles.}
To explore DataTailor addressing three principles for data selection, we clarify three principles based on the following insights: (a) \textbf{informativeness} is crucial for generalization when initializing MLLMs; (b) \textbf{uniqueness} is guaranteed to promote MLLMs after partial training by novel contributions; (c) \textbf{representativeness} improves relevant samples' performance during training. 
Specifically, we instantiate three settings to evaluate each of them essentially: 
(a) the improvement with 1\% data added to initial MLLMs reflects informativeness; (b) the improvement with extra 1\% data added to MLLMs based on 15\% random data training reflects uniqueness; (c) the average reduced loss of relevant samples with 1\% data added reflects representativeness.
We normalize each metric and compare it to three types of baselines in Fig.~\ref{test}(b).
The selected data from baselines show limitations in at least one principle. In contrast, DataTailor effectively selects valuable data for all three aspects of MLLM. Please refer to Appendix \textcolor{red}{D.1} for more details of instantiation analysis.

\noindent\textbf{Robustness Analysis of DataTailor.} 
(a) \textit{Model scale robustness.} In Table~\ref{parameter}, we verify the robustness of DataTailor with larger model scales. When keeping the same architecture and using LLaVA-v1.5-13B with higher parameters, our method demonstrates enhanced performance with larger scale MLLMs, especially +5.5 points on the LLaVA-Wild benchmark. This shows the stability of DataTailor at model scales. (b) \textit{Dataset robustness.} In Table~\ref{robust}, we explore the dataset robustness of DataTailor with two challenging perturbations of candidate datasets: \textbf{redundancy perturbation} and \textbf{noise perturbation}.
Specifically, we construct these perturbations with 50k redundant data and 50k noisy data by resampling and wrong answer combination. Note that we also sample 50k normal data from LLaVA-665k for balance. 
With redundancy and noise perturbations, we notice that DataTailor consistently demonstrates superior performance with limited data, whereas TIVE experiences a significant performance drop~(65.9 v.s. 54.1 of SciQA on redundancy disturbance and 48.5 v.s. 45.4 of GQA on noise disturbance).
It indicates DataTailor can bring out more distinctive and representative samples to identify the truly valuable samples from the redundant and noisy data for better robustness, which is crucial for discrimination tasks.

\begin{table}[!t]
\vspace{-0.2cm}
\setlength\tabcolsep{11pt}
\resizebox{0.48\textwidth}{!}{
\begin{tabular}{lcccccc}
\hline
\toprule
Methods & MMMU(val) & LLaVA-Wild & SciQA & GQA & Rel. \\ \hline
LLaVA-v1.5-13B \\
\quad\quad Full Data~(100\%) & 35.2 & 69.5 & 72.6 & 63.3 & 100.0\% \\
\quad\quad Random   ~(15\%) & 33.0 & 67.3 & 71.8 & 60.7 & 96.3\% \\
\quad\quad DataTailor~(15\%) & 36.4 & 75.0 & 73.9 & 61.2 & 102.4\% \\\hline
\end{tabular}}
\vspace{-0.3mm}
\caption{Model scale robustness analysis of DataTailor.}
\label{parameter}
\end{table}

\begin{figure}
  \centering
    \includegraphics[width=1.0\linewidth]{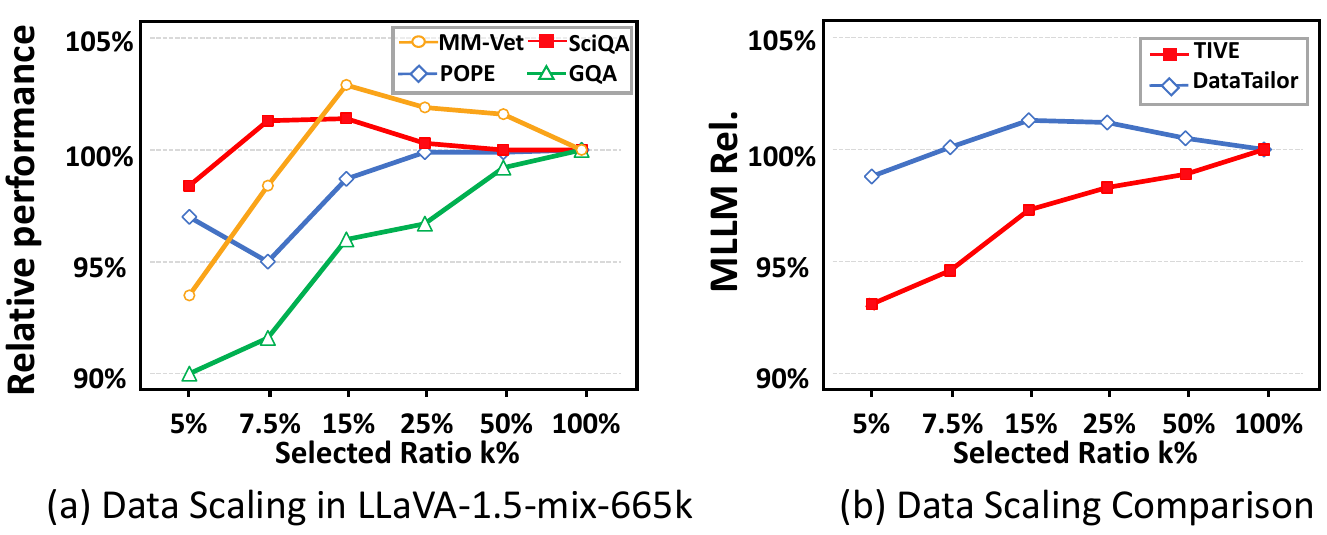}
     \vspace{-0.5cm}
  \caption{Ablation study of selection ratio $k\%$ in DataTailor.}
  \label{ablation_scale}
  \vspace{-0.3cm}
\end{figure}

\begin{table}[!t]
\vspace{-0.2cm}
\setlength\tabcolsep{11pt}
\resizebox{0.48\textwidth}{!}{
\begin{tabular}{lcccccc}
\hline
\toprule
\multicolumn{1}{c}{\multirow{2}{*}{Methods }} & 
\multicolumn{3}{c}{Redundancy Perturbation} & \multicolumn{3}{c}{Noise Perturbation} \\ 
& POPE & SciQA & GQA & POPE & SciQA & GQA \\ \hline
LLaVA-v1.5-7B \\
\quad\quad Full Data~(100\%) & 84.7 & 68.2 & 56.1 & 83.0 & 65.3 & 51.2\\
\quad\quad Random   ~(5\%) & 82.1 & 64.0 & 40.7 & 81.5 & 63.9 & 43.2\\
\quad\quad TIVE~(5\%) & 81.1 & 54.1 & 42.5 & 80.9 & 62.4 & 45.4 \\
\quad\quad DataTailor~(5\%) & 81.4 & 65.9 & 46.9 & 84.2 & 63.7 & 48.5\\\hline
\end{tabular}}
\vspace{-0.3mm}
\caption{Dataset robustness analysis of DataTailor with redundancy perturbation and noise perturbation.}
\label{robust}
\vspace{-0.5cm}
\end{table}
\noindent\textbf{Influence of Selection Proportion k\% in DataTailor.} As shown in Figure~\ref{ablation_scale} (a), when the selected data volume is relatively small, the model's performance improves as the data scale increases. However, due to the limited valuable data, further increasing the data volume introduces redundancy and noise, which degrades data-sensitive visual recognition performance~(\ie MM-Vet). This reveals the necessity of selecting data to ensure efficiency and maintain performance. 
Moreover, DataTailor quickly obtains over 100\% overall performance~(100.1\% with only 7.5\% data) while TIVE exhibits only limited performance while growing slowly, as shown in Figure~\ref{ablation_scale} (b). 
It indicates that DataTailor considers the uniqueness of the selected samples to avoid repetition and achieve continuous improvement.

\noindent\textbf{Computation Cost Analysis.} Since the overhead of data selection is crucial for effective pruning methods, we analyze the computational cost of DataTailor when selecting 7.5\% data. We find that TIVE even exceeds the original training cost, which has certain limitations. In contrast, DataTailor saves nearly 80\% of the overall time while outperforming the full model's performance. This confirms the efficiency of data selection with DataTailor for MLLMs.
 \begin{table}[!t]
 \vspace{-0.2cm}
    \centering
    \resizebox{0.48\textwidth}{!}{
    \begin{tabular}{l|cc|cc|cc}
        \hline
        &\multicolumn{2}{c|}{\textbf{Warmup}} & \multicolumn{2}{c|}{\textbf{Data Selection~(7.5\%)}} & \multicolumn{2}{c}{\textbf{Training}} \\ \hline
        & \textbf{Complexity} & \textbf{Actual} & \textbf{Complexity} & \textbf{Actual} & \textbf{Complexity} & \textbf{Actual} \\ \hline
        Full Model& - & - & - & - & $\mathcal{O}(|\mathcal{D}| \cdot |S|)$ & 100 H \\ \hline
        TIVE~\cite{liu2024less} & $\mathcal{O}(|\mathcal{D}| \cdot |S_{\text{warmup}|})$ & 8 H & $\mathcal{O}(|\mathcal{D}| \cdot |S|)$ & 100 H & $\mathcal{O}(|\mathcal{D}| \cdot |S^*|)$ & 7.5 H \\ 
        DataTailor~(Ours) & - & - & $\mathcal{O}(|S|)$ & 15 H & $\mathcal{O}(|\mathcal{D}| \cdot |S^*|)$ & 7.5 H \\         
        
        \hline
    \end{tabular}
    }\caption{Asymptotic complexity and wall-clock runtime (measured with 4*3090 for LLaVA-v1.5-7B experiments on LLaVA-mix-665k dataset). $|D|$ is the complexity of gradient computation.}\vspace{-0.3cm}
\end{table}

\begin{figure}[!t]
    \centering
    \includegraphics[width=1.\linewidth]{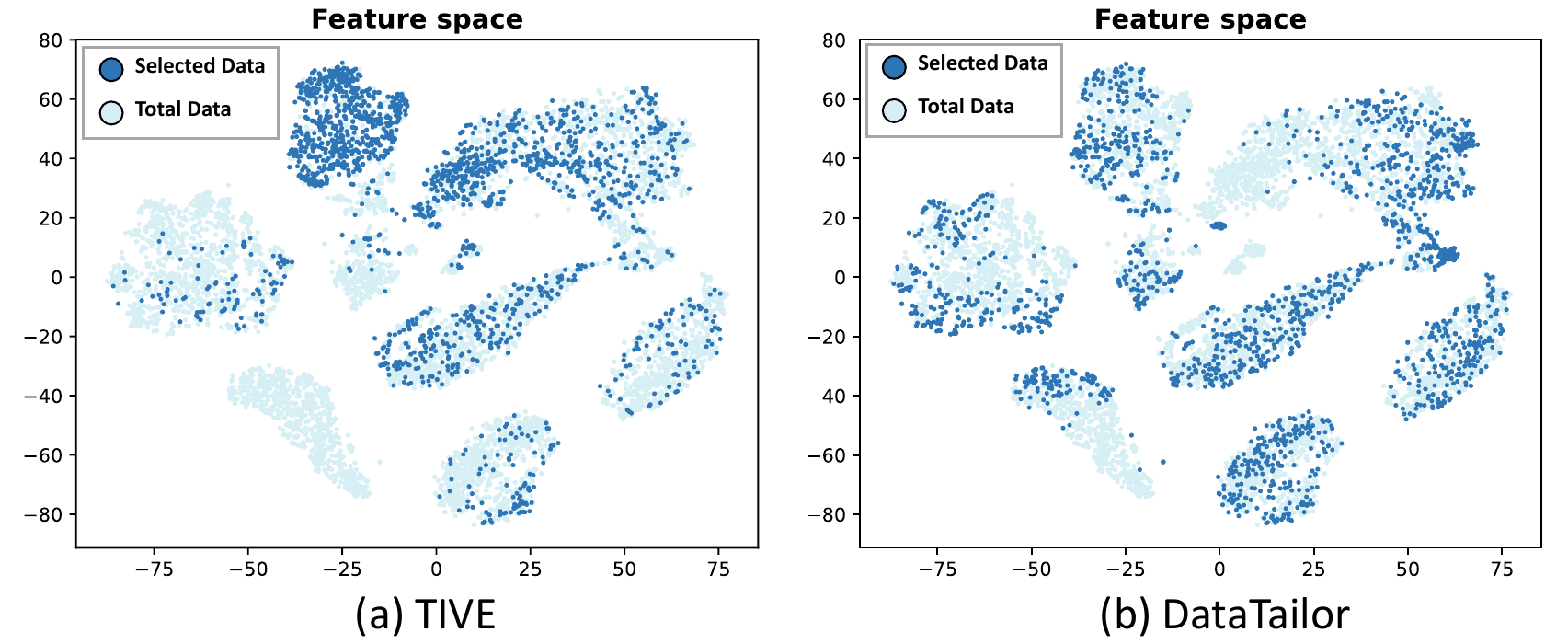}
   \vspace{-0.5cm}
    \caption{Visualization using t-SNE on feature space.} 
    \label{tsne}\vspace{-0.6cm}
\end{figure}

\noindent\textbf{Distribution of Selected Data.} To give an intuitive perspective on the selected data, we employ t-SNE~\cite{van2008visualizing} on the feature space of data chosen by DataTailor in Fig.~\ref{tsne}. Notably, DataTailor selects informative samples without redundancy or deviation, while TIVE, despite high informativeness, focuses solely on gradient similarity, leading to redundancy and outlier noise. This visualization confirms the effectiveness of our method adhering to three principles.


\section{Conclusion and Future Work}
In this paper, we reveal the drawbacks of existing data selection methods and identify three systematic principles of informativeness, uniqueness, and representativeness as fundamental to optimizing multi-modal data selection.
Building on this, we propose a unified framework, DataTailor, to synergistically integrate these principles for value evaluation and adaptively address the varying structure and complexity of samples across diverse tasks, thereby mastering collaborative multi-modal data selection.
Comprehensive experiments on the challenging MLLM and general VQA benchmarks show that DataTailor significantly improves the performance of MLLM data selection. In the future, we aim to extend DataTailor to more challenging interleaved datasets with extra modalities such as video and audio.

\noindent\textbf{Acknowledgment.} This work was supported by the National Natural Science Foundation of China (62436007), the Key R\&D Projects in Zhejiang Province (No. 2024C01106, 2025C01030), the Zhejiang NSF (LRG25F020001) and Wallenberg-NTU Presidential Postdoctoral Fellowship. We thank all the reviewers for their valuable comments.
{
    \small
    \bibliographystyle{ieeenat_fullname}
    \bibliography{main}
}
\clearpage
\appendix
\clearpage
\setcounter{page}{1}
\maketitlesupplementary
\setcounter{equation}{0}

\section{Overview}
In this supplementary material, we present:

\begin{itemize}
	\item More detailed analyses of DataTailor~(Section \ref{sec:b}).
	
	\item More experimental details~(Section \ref{sec:c}).
	
	\item Additional Experiment Analyses~(Section \ref{sec:d}).
\end{itemize}
\section{DataTailor Framework}
\label{sec:b}

\subsection{Principle Definition}
To avoid the computational infeasibility of the impact of samples on downstream task performance, we define three principles based on the geometric statistical properties of the samples to serve as practical approximation for assessing the value of multi-modal instructions:

\noindent\textbf{Definition B.1~(Informativeness).} It quantifies the samples' difficulty for downstream tasks within latent space. The sample difficulty is related to its feature distribution diversity. A sample $s\in S, S\subseteq D$ is high-informativeness if its token-level feature distribution is more diverse, indicating it contains a wider variety of information at the token level. By prioritizing such samples during training, models can learn more robust representations, thereby improving inference accuracy on downstream tasks.

Intuitively, simple samples often contain redundant information~(\eg, images with large areas of meaningless background or answers with repetitive descriptions). In such cases, the model can ``trickily" focus on a subset of information to complete the task. Due to the high similarity of tokens in the sample, the feature matrix columns (or rows) exhibit strong linear dependence, meaning the matrix has a low rank and contains significant redundancy. According to Singular Value Decomposition (SVD), smaller singular values decrease significantly, and larger ones become disproportionately larger, leading to lower singular value entropy. Mathematically, if the feature matrix $\mathbf{M}$ has low rank, its singular value matrix $\Sigma$ can be represented as:
\begin{equation}
\mathbf{M}=U\Sigma V
\end{equation}
where the smaller singular values in $\Sigma$ approach zero, while the larger singular values dominate. The singular value entropy (SVE) is defined as the entropy of normalized singular values~\cite{chen2023understanding}, which is computed as:
\begin{equation}
    \mathbf{SVE} = -\sum_{i=1}^r p(\sigma_i)\log p(\sigma_i)
\end{equation}
where $\sigma_i$ is a singular value, $p(\sigma_i)$ is the normalized probability distribution of the singular values, and $r$ is the rank of the singular value matrix. Since many of the singular values are close to zero, the entropy is low, reflecting the simplicity of the sample. In contrast, when the feature matrix of a sample is more information-rich and closer to full rank, singular values contribute more evenly, leading to higher entropy, which indicates a more complex sample. 

Therefore, singular value entropy serves as a practical approximation of sample informativeness, enabling the selection of more challenging samples that encapsulate a diverse range of information for downstream tasks.



\noindent\textbf{Definition B.2~(Uniqueness).} It measures deviation from local data density. A sample $s$ is high-uniqueness in neighborhood $\mathcal{N}(s)$ if:
\begin{equation}
    \min_{s' \in \mathcal{N}(s)} \|s - s'\| \geq \delta \cdot \text{diam}(\mathcal{N}(s))
\end{equation}
where $\delta\in(0,1)$ thresholds the relative margin and $\text{diam}(\cdot)$ is the diameter of the subset within the intra-cluster space. Due to local density constraints, models are forced to learn non-degenerate decision boundaries~\cite{devroye2013probabilistic}, which ensures the distinction of sample distributions and improves adversarial robustness. According to Lemma 2.4~\cite{balcan2008discriminative}, maximizing the Euclidean intra-cluster margins achieves this through geometric packing in $\ell_2$-space, which is equivalent to sampling along the data manifold boundary to reduce redundancy between samples thereby improving training robustness~\cite{learning2006semi}. It is worth noting that the dynamic computation based on the selected sample may be influenced by the greedy side effects during the selection process, making it difficult to achieve a global optimal solution.

Therefore, the static Euclidean distance between all neighborhood samples in the intra-cluster space can approximately capture data density of sample sets,  allowing the selection of unique samples for continual performance improvement. Given a sample $s$ and its neighbor samples $x_j\in\mathbf{C}$, the Euclidean distance $d_{i,j}$ can be calculated as follows:
\begin{equation}
    d_{i,j} = \Vert\mathbf{p_j}-\mathbf{p_i}\Vert_2
\end{equation}
where a smaller Euclidean distance of two samples in the latent space indicates that these samples are highly similar, leading to a lower uniqueness value.

\noindent\textbf{Definition B.3~(Representativeness).} It ensures samples adhere to population-level statistics. A sample $s$ is representative if its feature vector $\phi(s)$ satisfies:
\begin{equation}
    W_1(\mathcal{P}_S, \mathcal{P}_D) \leq \gamma
\end{equation}
where $W_1$ is Wasserstein-1 distance, $\mathcal{P}_S$ is average feature of the subset, and $\mathcal{P}_D$ is average feature of the global distribution. By incorporating the Wasserstein-1 constraint, it minimizes domain shift and stabilizes the dynamics of gradient descent~\cite{bottou2018optimization}. According to Proposition 1~\cite{paty2019subspace}, cosine similarity to cluster centroids approximates $W_1$-optimal transport under spherical normalization. Therefore, the similarity distribution from the inter-cluster space can approximately capture the alignment between samples and the overall data distribution for minimizing domain shift. The representativeness value can be measured by the cluster density of the sample, which is commonly estimated using the similarity distribution between cluster centroids:
\begin{equation}
    \tau_i^c=\frac{1}{K-1}\sum_{k\neq c}^K\exp({\operatorname{sim}}(\mathbf{\overline{p_k}},\mathbf{\overline{p_c}}))
\end{equation}
where $\mathbf{\overline{p_c}}$ is the feature of the cluster centroid that is calculated by the average feature of samples in the cluster. Specifically, high similarity indicates that the cluster containing the sample is well-aligned with other clusters, making it a strong representative of the overall data distribution.

\subsection{Adaptive Data Proportion for Data Selection}
To adapt to various task complexities within multi-modal datasets, we introduce adaptive proportion of selected data for each task. The challenge stems from the fact that task difficulty is inherently difficult to assess directly. We observe that data lacking directional diversity causes the generated trajectories to collapse into a limited subspace dominated by a few principal components. This reflects the complexity of the task, highlighting the need for more training data to enhance the robustness of MLLMs. Specifically, given the feature matrix $\mathbf{M}$ of a sample $s_i$, the largest singular value calculation relies on Singular Value Decomposition~(SVD):
\begin{equation}
    \mathbf{M} = U\Sigma V^T, \Sigma=\{\sigma_1,...,\sigma_r\}
\end{equation}
where $\Sigma$ is the singular value matrix and $r$ represents the singular value matrix rank of each sample. Here $r=L_i$ since the sample feature dimension $d$ is much larger than the sample token length $L_i$ and $\sigma_i$ represents each singular value.

\noindent\textbf{Definition B.4~(Largest Singular Value Ratio).} The largest singular value ratio~(LSVR) is defined as the ratio of the maximum singular value to the sum of singular values. LSVR captures the distribution of significant eigenvectors in a particular direction and reflects the task difficulty for training robustness.
\begin{equation}
    \mathbf{LSVR} = \frac{\sigma_{\text{max}}}{\sum_{j=1}^r \sigma_i}
\end{equation}
where $\sigma_{\text{max}}$ denotes the dominant singular value. When the LSVR becomes significantly large, it indicates insufficient variation in eigenvectors along the principal direction as follows: 
\begin{equation}
    \frac{\sigma_{\text{max}}}{\sum_{i=1}^r \sigma_i} \gg \frac{1}{r} 
\end{equation}

This indicates that information is concentrated in specific singular value directions in the latent space. Such spectral imbalance in singular values reduces the effective dimensionality of learned representations, meaning that less knowledge can be extracted from individual samples. Consequently, such difficult tasks with larger LSVR require higher data selection proportions to ensure sufficient learning. Specifically, we compute the average of the LSVR for all samples in the task as follows:
\begin{equation}
\quad x_p = avg({\frac{\sigma_{max}}{\sum_{j=1}^{r}\sigma_j}})
\end{equation}
To amplify the contribution of task difficulty to data selection, we square the average largest singular value ratio and normalize it based on the number of samples corresponding to each task, yielding the data selection ratio as follows:
\begin{equation}
k_p=\frac{x_p^2\cdot |S_p|}{\sum_{q}x_q^2\cdot |S_q|}\cdot k
\end{equation}
where $|S_q|$ is the corresponding sample number of each task. Then, we adjust the data selection rate of each task from $k$ to $k_p$ to achieve task-adaptive proportions. Once the data selection ratio for each task is determined, we utilize the synergistic sample value from DataTailor to perform collaborative multi-modal data selection for each task.
\subsection{Cross-Modal Domain Clustering}

\subsubsection{Algorithmic Formulation}
The clustering pipeline commences with inter-sample affinity quantification via $\ell_2$-norm distance measurements within each initial category. Starting with all nodes as individual clusters, we iteratively merge the pair (A,B) exhibiting the minimal increase in Sum of Squared Errors (SSE) to construct a dendrogram:

\begin{equation}
A, B = \mathop{\arg\min}\limits_{A,B \in \mathcal{P}} \Delta \mathrm{SSE}(A, B)
\end{equation}
where $\mathcal{P}$ denotes the partition of nodes into distinct clusters. The SSE increment from merging clusters $A$ and $B$ is defined as:

\begin{equation}
    \Delta \mathrm{SSE}(A,B) = \frac{n_A n_B}{n_A + n_B} \|\boldsymbol{\mu}_A - \boldsymbol{\mu}_B\|_2
\end{equation}
with $n_A, n_B$ representing cluster sizes and $\boldsymbol{\mu}_A, \boldsymbol{\mu}_B$ their centroids. This criterion minimizes intra-cluster variance growth.
To theoretically characterize the merging behavior, we formalize a key monotonicity property inherent to Ward's algorithm.

\begin{theorem}[SSE Monotonicity in Ward's Method]\label{thm:sse}
For a hierarchical merging process $\{\mathcal{P}_k\}_{k=0}^{n-1}$ under Ward's algorithm (where $\mathcal{P}_0$ contains singleton clusters), the total SSE increment satisfies the non-decreasing property:
\[
\Delta \mathrm{SSE}(\mathcal{P}_k) \geq \Delta \mathrm{SSE}(\mathcal{P}_{k-1}), \quad \forall k \geq 1,
\]
where $\Delta \mathrm{SSE}(\mathcal{P}_k)$ denotes the incremental SSE from merging $\mathcal{P}_{k-1}$ to $\mathcal{P}_k$.
\end{theorem}

This monotonic progression ensures that earlier merges correspond to more natural cluster unions, while later merges sacrifice increasing amounts of variance. Leveraging this property, we define a threshold-based partitioning rule to extract clusters from the dendrogram hierarchy. A dendrogram \( T = (V, E) \) consists of leaf nodes \( V_{\text{leaf}} = \{s_1, ..., s_n\} \) and internal nodes, which serve as merge points annotated with \( \Delta\mathrm{SSE} \) values. Given a threshold \( T = \lambda \cdot \Delta\mathrm{SSE}(\mathcal{P}_{n-1}) \), the optimal partition is determined by:
\[
\mathcal{P}^* = \max\left\{k \mid \Delta\mathrm{SSE}(\mathcal{P}_k) \leq T\right\}
\]

\subsubsection{GPU-Accelerated Distance Computation}
For efficient pairwise distance measurement between $n$ samples in $\mathbb{R}^d$, we implement a parallelized $\ell_2$-norm computation framework using CUDA-optimized matrix operations. The distance matrix $D \in \mathbb{R}^{n \times n}$ can be derived through the algebraic identity, which:


\begin{equation}
D_{x,y} = \sqrt{S_{x,x} + S_{y,y} - 2 S_{x,y}}
\end{equation}

where $X \in \mathbb{R}^{n\times d}$ represents the feature matrix containing $n$ samples, and $S = XX^T \in \mathbb{R}^{n\times n}$ represents the similarity matrix. 
Our kernel-based implementation achieves $\mathcal{O}(n^2d/m)$ theoretical speedup through massive parallelization across GPU cores, effectively transforming an originally $\mathcal{O}(n^2d)$ complexity operation into a highly parallelizable matrix multiplication task. In addition, we use the parallel pipeline strategy to extract feature vectors of the feature matrix while calculating the $\ell_2$-norm distance between sample features for efficiency.

\subsubsection{Optimized Hierarchical Clustering Merge}
To address the quadratic complexity inherent in conventional hierarchical clustering, we implement a memory-efficient variant of the nearest-neighbor chain algorithm. This optimization framework features:
\begin{itemize}
    \item Randomized stack initialization with cluster prototypes
    \item Iterative nearest-pair identification via stack
    \item In-stack merging with $\mathcal{O}(n)$ per-operation cost, where each node undergoes $\mathcal{O}(1)$ amortized operations.
\end{itemize}

The proposed acceleration strategy reduces computational overhead to $\mathcal{O}(n^2)$ while preserving the theoretical guarantees of Ward's method.  By optimizing the clustering process, the time consumption of the clustering was accelerated by $90\times$, resulting in significantly improved performance.



\begin{table}[!t]
\renewcommand\arraystretch{1.5}
\resizebox{0.475\textwidth}{!}{
\begin{tabular}{c|ccccccc}
\hline
Threshold $\lambda$ & w/o clustering & 0.05 & 0.1 & 0.25 & 0.5  \\ \hline
MLLM Rel.& 98.1\% & 99.8\% & 101.3\% & 100.0\% & 98.5\%\\\hline
\end{tabular}
}
\caption{The analysis of different similarity thresholds for cross-modal domain clustering in extrinsic value estimation.}
\label{lambda}
\vspace{-0.4cm}
\end{table}
\subsubsection{Threshold Parameter Analysis}
Since the quality of clustering is critical for the domain-based adaptive data proportion in DataTailor, we further explore the effect of dynamic threshold to cross-modal domain clustering on the multi-modal data selection in Table~\ref{lambda}. The clustering threshold determines the cluster size based on the dataset's sample distribution. Therefore, setting it close to the overall data selection proportion ensures an appropriate size of clusters that effectively captures sample relationships.
Our experiments reveal that low or high thresholds compromise the constraints on the uniqueness or representativeness of high-quality samples, leading to lower performance of DataTailor. Thus, we set the appropriate threshold $\lambda$ as $0.1$ for cross-modal domain clustering, which is close to the total data selection proportion.
\subsection{Balance between Three Principles}
Since multi-modal samples exhibit varying structures, we propose an adaptive weight to combine the three principal values. We restate the underlying principles behind the three properties to show their effectiveness: (a) Informativeness. It determines external relationships due to its core training contributions. Because MLLMs rely on token-level inputs, the SVD of token feature space reveals the samples’ contribution to the MLLM. That's why we use singular value entropy to reflect the value of samples for generalization. (b) Uniqueness. For repeated samples, their uniqueness is adjusted during selection based on the distribution of chosen points by normalization, ensuring duplicates are treated differently. 
(c) Representativeness. It aims to isolate undesired noisy samples that may have a high uniqueness score, while general noisy data can be identified by its lower informativeness (i.e., average 0.297 in clusters).

In general, it is important to balance the above three principles to select data collaboratively. On the one hand, we explore various ratios of these two external values~(\ie, uniqueness and representativeness) in Figure~\ref{balance}. The smooth performance transition~(less than 2\%) near the 1:1 ratio suggests that the trade-off between them remains stable. 
Therefore, we select the 1:1 ratio as the optimal value for collaboration. On the other hand, we use adaptive weights between the information values and the two values for the varying instruction rounds of the samples in the dataset.
\begin{table}[!t]
\renewcommand\arraystretch{1.5}
\resizebox{0.475\textwidth}{!}{
\begin{tabular}{c|ccccc}
\hline
$V^{Uni}:V^{Rep}$ & 0:1 & 0.5:1 & 1:1 & 1:0.5 & 1:0  \\ \hline
MLLM Rel.& 99.6\% & 100.5\% & 101.3\% & 100.3\% & 99.5\%\\\hline
\end{tabular}
}
\caption{Balance between uniqueness and representation in DataTailor for data selection of MLLMs.}\label{balance}
\vspace{-0.4cm}
\end{table}


    

\section{More Experimental Details}
\label{sec:c}
\subsection{Implementation Details} 
Following prior research~\cite{wei2023instructiongpt, liu2024less} and each dataset scale, we keep 5\% as the data proportion~(0.2k) for data selection on MiniGPT4-Instruction~\cite{zhu2023minigpt} and 7.5\% as the data proportion~(50.0k) for data selection on on LLaVA-1.5-mix-665k~\cite{liu2024improved} for the standard setting. In the transferability analysis, we uniformly set 5\% as the data proportion~(12.3k) for data selection on mPLUG-Owl-7B-264k-Instructions~\cite{ye2024mplug} and 5\% as the data proportion~(34.7k) for data selection on Bunny-695k~\cite{he2024efficient}. During the data selection process, we retain all parameters from the original model but freeze all gradients. DataTailor evaluates the values of the three principles for multi-modal samples using the initialized features of the pre-trained model. This allows DataTailor to select high-quality samples while efficiently maintaining strong transferability.

During data selection in DataTailor, we balance the uniqueness and representativeness of different clusters.  First, we normalize uniqueness and representativeness values across clusters by removing the influence of spatial distribution and cluster size. Next, we standardize the impact of the average sample value across clusters. Specifically, within the same task, we uniformly scale informativeness, uniqueness, and representativeness metrics to a $[0, 1]$ range to ensure consistent distribution alignment. By harmonizing these normalized values across samples, we enable collaboration among Informativeness, Uniqueness, and Representativeness values.  This methodology fosters balanced metric collaboration in our adaptive data selection framework, ensuring proportional consideration of all three criteria.

During fine-tuning, we apply the LoRA strategy~\cite{hu2021lora} to fine-tune each dataset and its subsets from various data selection methods due to the limited GPU resources. For LLaVA-v1.5-7B, we use 4*3090 GPUs for fine-tuning, where the batch size of each device is set to 12 and the training epoch is set to one epoch. For MiniGPT-4-7B, we use 1*A6000 GPU for fine-tuning, where the batch size of each device is set to 12 and the training epoch is set to 5 epoch. During fine-tuning, we only distinguish the dataset scale through various data selection methods and keep all other training parameters consistent for a fair comparison. 
\begin{table}[!t]
\resizebox{0.48\textwidth}{!}{
\begin{tabular}{l|ccc|c}
\hline
\toprule
\multirow{2}{*}{Methods} &\multicolumn{3}{c}{Selected Data Evaluation}&\multirow{2}{*}{MLLM Rel.}\\
& \multicolumn{1}{c}{Informativeness} & \multicolumn{1}{c}{Uniqueness} & \multicolumn{1}{c}{Representativeness} &  \\\hline
IFD~(7.5\%)  & \textcolor{black}{32.3}&\textcolor{black}{0.341}&\textcolor{black}{30.3} & 87.3\%\\
INSTAG~(7.5\%) & \textcolor{black}{30.9}&\textcolor{black}{0.347}&\textcolor{black}{34.4} & 96.4\% \\
LESS~(7.5\%)  & \textcolor{black}{34.0}&\textcolor{black}{0.314}&\textcolor{black}{33.3} & 94.3\%\\\hline
\quad +$V_i^{Inf}$~(7.5\%) & \textcolor{black}{34.5}&\textcolor{black}{0.348}&\textcolor{black}{34.8}&98.0\%\\
\quad +$V_i^{Uni}$~(7.5\%) & \textcolor{black}{33.4}&\textcolor{black}{\textbf{0.364}}&\textcolor{black}{34.5}&97.3\% \\
\quad +$V_i^{Rep}$~(7.5\%) & \textcolor{black}{33.9}&\textcolor{black}{0.343}&\textcolor{black}{\textbf{35.0}}&97.5\%\\
\textbf{DataTailor}~(7.5\%) &\textcolor{black}{\textbf{34.8}} &\textcolor{black}{0.358}&\textcolor{black}{34.9}& 100.1\%\\
\bottomrule
\end{tabular}
}
\caption{Quantitative Valuation of Three Principles. All setups are on the model of LLaVA-v1.5-7B and the dataset of LLaVA-mix-665K for both data selection and MLLM training.}\label{threeprinciples}
\end{table} 

\subsection{Candidate Datasets Details}
\noindent\textbf{MiniGPT4-Instruction.}
It contains approximately 3,500 instruction pairs, each consisting of an image and a corresponding detailed description. The correctness of each image description is manually verified to ensure high quality.

\noindent\textbf{LLaVA-v1.5-mix-665k.} This is currently the most extensive multimodal instruction dataset, encompassing instruction data across a wide range of tasks. It contains a variety of datasets: VQA~\cite{antol2015vqa}, OCR~\cite{mishra2019ocr}, region-level VQA~\cite{krishna2017visual}, visual conversation~\cite{liu2024visual} and language conversation~\cite{ShareGPT} data. For all datasets, QA pairs from the same training image are merged into a single conversation, and excessively long data is filtered out to improve training efficiency. As a result, this process yields 665k instruction pairs across 13 tasks.

\noindent\textbf{mPLUG-Owl-7B-264k-Instructions.} It gathers pure text instruction data from two distinct sources: 52k data from the Alpaca~\cite{taori2023stanford} and 54k from the Baize~\cite{xu2023baize}. Additionally, it involves 158k multi-modal instruction data from visual conversations in the LLaVA dataset~\cite{liu2024visual}. In this way, it incorporates both pure text instruction data and multimodal instruction data, demonstrating that DataTailor is well-suited for diverse data selection tasks.

\noindent\textbf{Bunny-695k.} It primarily utilizes SVIT-mix-665k~\cite{zhao2023svit}, replacing ShareGPT-40k~\cite{ShareGPT} with WizardLM-evol-instruct-70k~\cite{xu2024wizardlm} to create Bunny-695k. Compared to LLaVA-665K, this dataset contains more complex multi-modal instructions, enabling the evaluation of DataTailor’s ability to transfer to more intricate multi-modal data selection.

\begin{figure}[!t]

  \centering

   \includegraphics[width=1.\linewidth]{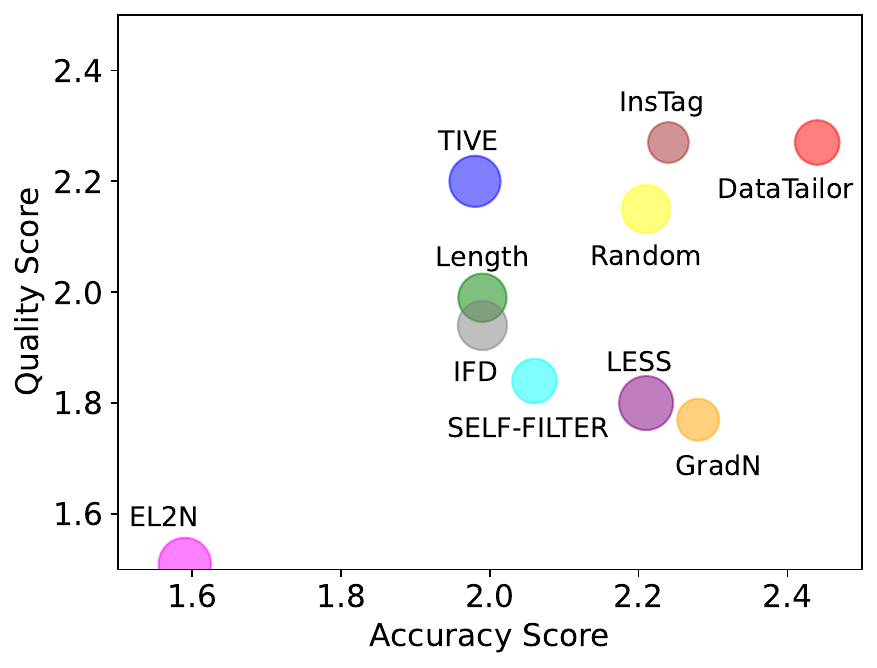}
    
  \caption{Quality score~(y-axis, higher is better), accuracy score~(x-axis, higher is better), and the stability~(circle sizes, smaller is better) of MLLMs' responses on OwlEval benchmark. We set the data selection ratio for each method to 7.5\%.}\label{owleval}
  \label{owleval}
\end{figure}

\section{Additional Experiment Analyses}\label{sec:d}
\subsection{Quantitative Valuation of Three Principles}
To quantitatively investigate how DataTailor addresses each of the three principles for data selection, we designed three experimental settings to assess them. Here we show the detailed metric values for DataTailor and other baselines on three experimental setups in Table~\ref{threeprinciples}. 

\begin{table}[!t]
\resizebox{0.475\textwidth}{!}{
\begin{tabular}{lccccc}
\hline
\toprule
\multirow{1}{*}{Methods} & Dataset & MMMU&
MathVerse & MathVision&\multirow{1}{*}{MLLM \textcolor{red}{Avg.}}\\\hline
QVQ-72B-preview  & - & 66.0 & 41.5 & 38.2 & 48.6 \\
InternVL2.5-78B  & - & 70.0 & 39.2 & 32.2 & 47.1 \\\hline
Qwen2-VL-7B & Virgo(100\%) & 46.7 & 36.7 & 24.0 & 35.8 \\
\quad w/ random & Virgo(15\%) & 46.7 & 32.5 & 23.4 & 34.2  \\
\quad w/ DataTailor & Virgo(15\%) & 50.1 & 34.8 & 23.8 & \textbf{36.2} \\\hline
Qwen2-VL-72B & Virgo(100\%) &  65.0 & 48.1 & 38.6 & 50.6 \\
\quad w/ random & Virgo(15\%) & 59.4 & 43.7 & 36.8 & 46.6 \\
\quad w/ DataTailor & Virgo(15\%) & 63.0 & 46.2 & 40.3 & \textbf{49.8} \\
\bottomrule
\end{tabular}
}
\caption{The scalable results on math and reasoning benchmarks of DataTailor with more scalable MLLMs.}
\label{qwen2.5}
\end{table}

\subsection{More scalable MLLMs Results}
Further, we apply DataTailor to \textbf{larger and newer backbones}~(\textit{i.e.}, Qwen-2-VL-7B \& 72B) for more robust evaluation in the Table~\ref{qwen2.5}.
Since Qwen's data is closed-source, we fine-tune on the open Virgo dataset~\cite{du2025virgo}. 
Similarly, we observe that DataTailor exhibits competitive performance with only 15\% data~(36.2 v.s. 35.8 of full data), outperforming other \textbf{open-source MLLMs} baselines.

\subsection{Instantiation Comparisons of Three Principles}
\begin{figure*}[!t]

  \centering

   \includegraphics[width=1.\linewidth]{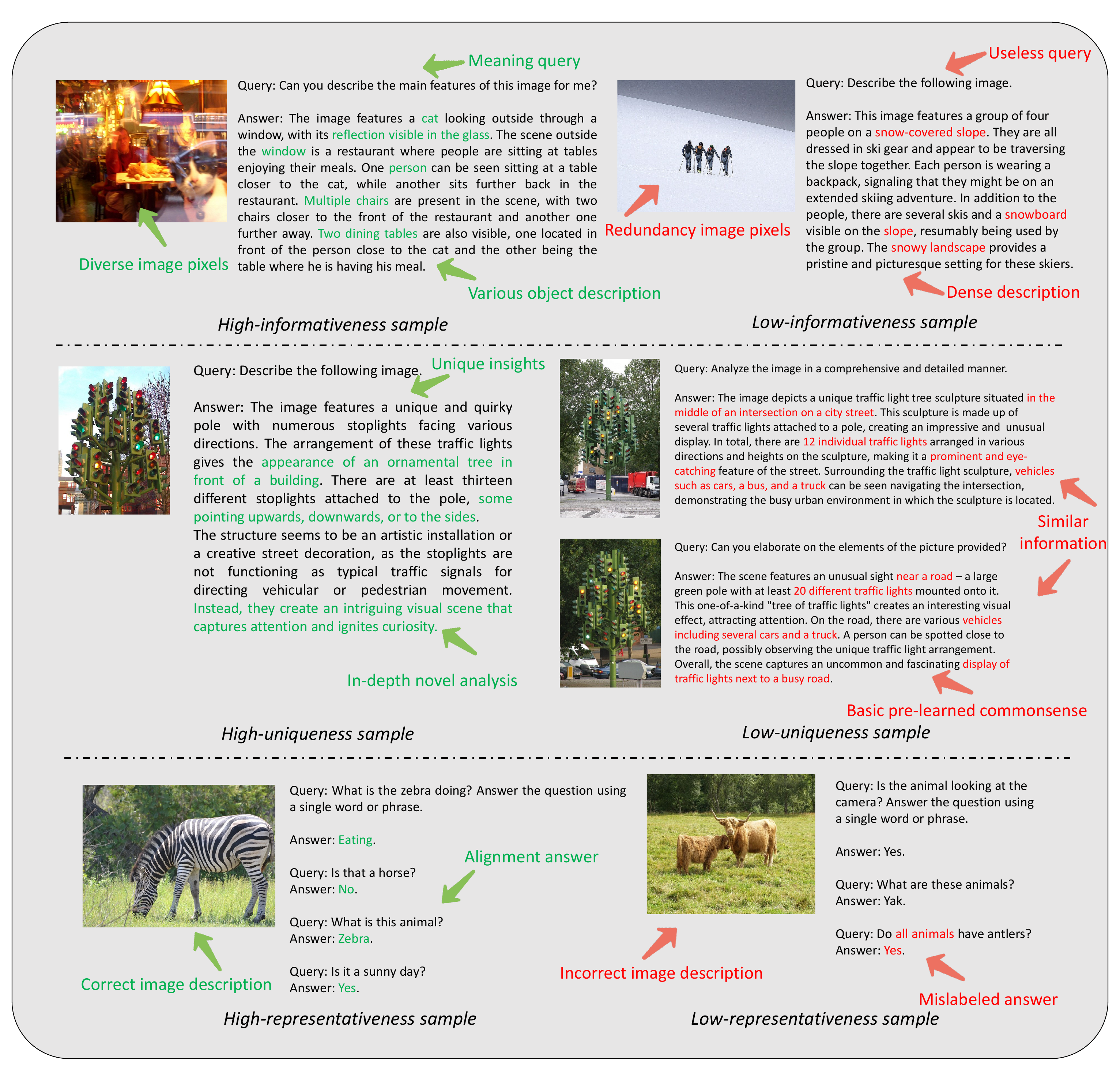}
    
  \caption{Instantiation Comparisons of DataTailor in addressing three core principles~(\ie informativeness, uniqueness, and representativeness) for selecting valuable multi-modal instruction samples.}
  \label{example}
\end{figure*}

To fully explore the characteristics of valuable samples that are meaningful for downstream tasks, we further present a few instantiation comparisons of sample characteristics from the three principal perspectives in Figure~\ref{example}. 
The left side shows the data preferred by DataTailor, and the right side shows the remaining data.

From the perspective of \textbf{informativeness}, we can observe that the samples selected by DataTailor contain richer information and various description, whereas other samples suffer from excessive redundancy in responses and numerous blank pixels in images~(\eg, the right image shows only snow slope and the text repeating snowboard). This indicates that selecting samples according to informativeness can select samples as complex as possible to improve inference accuracy in downstream tasks while also facilitating the generation of richer and higher-quality content.

From the perspective of \textbf{uniqueness}, we can observe that the samples selected by DataTailor in each cluster contain unique insights~(\eg, critically mention artistic installation) and in-depth novel analysis~(\eg, analyze diversity of the ornamental tree). However, other examples in the cluster show similar information and primarily describe basic, pre-learned commonsense knowledge, which limits MLLMs' ability to enhance generalization on downstream tasks continuously. This suggests that selecting samples based on uniqueness enables the inclusion of distinctive samples, which provide deeper and more novel insights to continuously enhance the reasoning capability during the fine-tuning phase of MLLM.

From the perspective of \textbf{representativeness}, we observe that the samples selected by DataTailor exhibit accurate descriptions and aligned answers. This is because the method ensures that the selected samples represent the overall distribution, avoiding noise and mislabeled data. Samples with low representativeness tend to exhibit mislabeled answers and outlier features, which can lead to incorrect optimization directions, ultimately hindering the performance improvement of downstream tasks. This suggests the need for assessing the representativeness of samples to filter out noisy or mislabeled data.

\begin{table}[!t]
\vspace{-3.5mm}
\resizebox{0.475\textwidth}{!}{
\begin{tabular}{lccccc}
\hline
\toprule
\multirow{1}{*}{Methods} &\multicolumn{1}{c}{Informativeness} & \multicolumn{1}{c}{Uniqueness} & \multicolumn{1}{c}{Representativeness}&\multirow{1}{*}{MLLM Rel.}\\\hline
+$V^{Inf}$~(15\%)& 34.3(\textcolor{magenta}{33.1})& -& -&99.2\%(\textcolor{magenta}{97.7\%})\\\hline
+$V^{Uni}$~(15\%)& -& 0.363(\textcolor{magenta}{0.347})& -&98.5\%(\textcolor{magenta}{97.4\%})\\\hline
+$V^{Rep}$~(15\%)& -& -& 34.6(\textcolor{magenta}{32.5})&98.6\%(\textcolor{magenta}{97.6\%})\\
\bottomrule
\end{tabular}
}
\caption{Deep Analysis for Calculation of Three Values}
\label{three_ablation}
\end{table}
Based on the above, we propose three essential principles~(\ie, informativeness, uniqueness, and representativeness) for the practical selection of valuable samples that truly contribute to the downstream inference performance of MLLMs. 
To further illustrate the superiority of these three quantitative metrics for three principles, we explore the insights underlying three values and conduct more \textbf{ablation} of potential alternatives~(results are of \textcolor{magenta}{red color} in Table~\ref{three_ablation})
For \textbf{informativeness}, we adopt the SVE based on the theory that more informative samples have feature matrices closer to full rank, leading to more uniform singular values and thus higher SVE. 
We explore pair-wisely calculating mutual information between token-level features as an alternative.
As SVE captures the overall distribution of feature directions, it achieves better informativeness and MLLM performance for data selection~(\textcolor{magenta}{99.2\%} v.s. \textcolor{magenta}{97.7\%}).
For \textbf{Uniqueness}, we adopt the Euclidean distance based on the theory that larger distances in the feature space separate more unique samples and samples containing pre-learned common knowledge are clustered near the center. We explore directly selecting one sample from each cluster as an alternative. As Euclidean distance measures samples' deviation from dense regions, it obtains better MLLM performance with uniqueness~(\textcolor{magenta}{98.5\%} v.s. \textcolor{magenta}{97.4\%}).
For \textbf{Representativeness}, we adopt cosine similarity based on the theory that it reflects the true directional alignment of the overall distribution.
We explore using the overall Euclidean distance between samples as an alternative. 
As cosine similarity emphasizes overall directional alignment instead of being influenced by the magnitude of outlier samples like distance, it performs better in MLLM with true representativeness~(\textcolor{magenta}{98.6\%} v.s. \textcolor{magenta}{97.6\%}).


\subsection{Limitations}
We observe some failure cases that DataTailor cannot discriminate long-term instructions when the reasoning process is omitted, \textit{e.g.}, math reasoning and code program.
\subsection{Human Evaluation}
To comprehensively evaluate whether data selection ensures the open-ended capabilities of MLLMs, we conduct further human evaluations using the OwlEval benchmark. OwlEval~\cite{ye2024mplug} is an open-ended evaluation set comprising 82 artificially constructed questions. We evaluated responses from all models on a 3-0 scale (aligned with option A-D in the official setting), assessing quality based on informativeness and alignment with the question, and accuracy based on consistency with image content. Furthermore, we calculate the score variance for all responses of the MLLMs using different data selection methods to assess model stability. We visualize the human-evaluation results in Figure~\ref{owleval}. 
We observe that using DataTailor for data selection best preserves the response capabilities of MLLMs, enabling them to provide both informative answers and maintain the highest level of accuracy. This demonstrates that DataTailor effectively selects representative samples to support the overall capabilities of MLLMs, addressing the challenge of collaborative multimodal data selection without overemphasizing specific abilities.

\end{document}